%% file: main.tex
\documentclass[10pt]{article}

\usepackage[letterpaper,top=1in,bottom=1in,left=1in,right=1in]{geometry}
\usepackage[T1]{fontenc}
\usepackage{lmodern}  
\usepackage{mathptmx}
\usepackage[scaled=0.90]{helvet}
\usepackage{microtype}
\usepackage{amsmath,amssymb,bm,mathtools}
\usepackage{booktabs,multirow,tabularx,colortbl,makecell,threeparttable}
\usepackage{enumitem}
\usepackage{xspace}
\usepackage{siunitx}
\usepackage{xcolor}
\usepackage{graphicx}
\usepackage{tikz}
\usepackage{pgfplots}
\pgfplotsset{compat=1.18}
\usepgfplotslibrary{groupplots}
\usetikzlibrary{arrows.meta,positioning,fit,backgrounds,calc,shapes.misc,shapes.geometric,decorations.pathreplacing,patterns}
\usepackage[font=small,labelfont=bf,skip=5pt]{caption}
\usepackage{subcaption}

\usepackage[numbers,sort&compress]{natbib}

\usepackage[colorlinks=true,linkcolor=oursblue,citecolor=oursblue,urlcolor=oursblue]{hyperref}
\usepackage[capitalise,noabbrev]{cleveref}

\usepackage{fancyhdr}
\input{figures/style}
\input{macros}

\begin{document}

\input{sections/00_title}
\input{sections/00_abstract}

\input{figures/fig_teaser}

\input{sections/01_introduction}
\input{sections/02_background}
\input{sections/03_method}
\input{sections/04_experiments}
\input{sections/05_transfer}
\input{sections/06_quantum}
\input{sections/07_related}
\input{sections/08_limitations}
\input{sections/09_conclusion}

\bibliographystyle{abbrvnat}
\bibliography{refs}

\clearpage
\appendix
\input{sections/A_implementation}
\input{sections/B_hardware}
\input{sections/C_extensions}

\end{document}

%% file: figures/style.tex
\definecolor{oursblue}{HTML}{1F5AA6}   
\definecolor{oursfill}{HTML}{E8F0FA}
\definecolor{agentorange}{HTML}{D2601F} 
\definecolor{agentfill}{HTML}{FCEEE4}
\definecolor{basegray}{HTML}{8B919A}   
\definecolor{basefill}{HTML}{F2F3F5}
\definecolor{qteal}{HTML}{0D7A6A}      
\definecolor{qfill}{HTML}{E2F3EF}
\definecolor{ink}{HTML}{30353D}
\definecolor{softink}{HTML}{6B717B}
\definecolor{rule}{HTML}{CDD1D7}
\definecolor{loopc}{HTML}{D2601F}      
\definecolor{prodc}{HTML}{1F5AA6}      
\definecolor{endc}{HTML}{0D7A6A}       

\tikzset{
  every picture/.append style={font=\sffamily},
  panel/.style={draw=#1, rounded corners=4pt, line width=0.7pt, fill=white},
  panelhead/.style={font=\sffamily\bfseries\footnotesize, text=#1, anchor=north west, inner sep=0pt},
  panelsub/.style={font=\sffamily\scriptsize, text=softink, anchor=north west, inner sep=0pt},
  card/.style={draw=#1, fill=#1!7, rounded corners=2.5pt, line width=0.5pt, align=center,
               font=\sffamily\scriptsize, inner xsep=4pt, inner ysep=3pt, text=ink},
  tok/.style={draw=#1, fill=#1!14, rounded corners=1.2pt, line width=0.4pt, minimum width=0.50cm,
              minimum height=0.34cm, inner sep=0pt, font=\sffamily\tiny, text=ink},
  arr/.style={-{Stealth[length=4.2pt,width=3.4pt]}, line width=0.65pt, draw=ink},
  darr/.style={{Stealth[length=4.2pt,width=3.4pt]}-{Stealth[length=4.2pt,width=3.4pt]}, line width=0.65pt, draw=ink},
  stat/.style={font=\sffamily\scriptsize, text=ink, align=center},
  qnode/.style={circle, draw=#1, fill=#1!12, line width=0.5pt, minimum size=0.34cm, inner sep=0pt},
}

\pgfplotsset{
  paperaxis/.style={
    axis line style={rule}, tick style={rule}, major tick length=2pt,
    tick label style={font=\sffamily\scriptsize, text=ink},
    label style={font=\sffamily\scriptsize, text=ink},
    title style={font=\sffamily\bfseries\scriptsize, text=ink, yshift=-3pt, align=center},
    legend style={font=\sffamily\scriptsize, draw=none, fill=none},
    every axis plot/.append style={line width=0.5pt},
    ymajorgrids, grid style={rule!60, line width=0.3pt},
    axis x line*=bottom, axis y line*=left, enlarge x limits=0.35,
    /pgf/number format/assume math mode=true, minor tick style={draw=none},
  },
  minibar/.style={paperaxis, ybar, /pgf/bar width=13pt, ymin=0, xtick=data, xticklabel style={font=\sffamily\scriptsize},
    nodes near coords, every node near coord/.append style={font=\sffamily\scriptsize\bfseries, text=ink},
    height=3.15cm, width=4.40cm, scaled y ticks=false},
}

%% file: macros.tex
\newcommand{\ours}{\textsc{Tree-Ising}\xspace}

\newcommand{\R}{\mathcal{R}}
\newcommand{\Pset}{\mathcal{P}}
\newcommand{\Zset}{\mathcal{Z}}
\newcommand{\Mset}{\mathcal{M}}
\newcommand{\Fset}{\mathcal{F}}
\newcommand{\E}{\mathbb{E}}
\newcommand{\dose}{\operatorname{dose}}
\newcommand{\coll}{\operatorname{coll}}
\newcommand{\uloop}{u_{\mathrm{loop}}}
\newcommand{\uend}{u_{\mathrm{end}}}
\newcommand{\code}[1]{\texttt{#1}}

\setlist{itemsep=2pt,topsep=3pt,leftmargin=1.4em}
\renewcommand{\arraystretch}{1.08}
\newcolumntype{Y}{>{\raggedright\arraybackslash}X}
\newcommand{\oursrow}{\rowcolor{oursfill}}

%% file: sections/00_title.tex
\thispagestyle{plain}
\begin{center}
  {\LARGE\bfseries Tree-Ising: Repairing Language Models\\[2pt] with Bit-Level Ising Edits\par}
  \vspace{12pt}
  {\normalsize Hadi Salloum\textsuperscript{1} \qquad Ammar Ali\textsuperscript{2}\par}
  \vspace{4pt}
  {\small \textsuperscript{1}QDeep, Riyadh, Saudi Arabia \qquad
   \textsuperscript{2}LEAD (Lab for Efficient AI and Data), ITMO University\par}
\end{center}
\vspace{2pt}

%% file: sections/00_abstract.tex
\begin{abstract}
Weight-space programming, introduced in recent MIT work by Takehi et al., enables a frontier LLM agent to read a network's weights and write a repair. Their GPT-6 Astra agent corrected the reasoning loops of Qwen3.5-0.8B without gradients, utilizing 158 agent turns, 89.7 million tokens, and 50 forward-pass requests. Our method, \ours, casts this repair as a sparse discrete optimization problem over the model's native BF16 weights. We leverage forward statistics to define an objective that weakens a loop direction, strengthens a concluding direction, and preserves the layer's output on productive tokens. Within a state-tree representation, where lossless BF16 storage acts as a special case, the admissible edits consist of sign flips and exponent shifts that preserve all shared group exponents and mantissas. We decompose this objective exactly into independent QUBO (Ising) problems, one per output row, which we solve on a GPU using simulated annealing or the quantum-inspired simulated bifurcation method. Evaluated on the two layers targeted by the agent, \ours achieves 54.6\% on MATH-500, outperforming the agent's edit (51.7\%) under the same evaluation protocol, while the base model scores only 7.9\%. Furthermore, the simulated bifurcation solver yields a statistically indistinguishable repair at 55.1\%. Crucially, both solvers outperform the agent edit across all five math benchmarks, with statistically significant improvements on MATH-500 and GSM8K after correcting for multiple comparisons. Moreover, our approach modifies 13\% to 17\% fewer weights and reduces BF16 bit flips by a factor of 2.2 to 2.3. It requires only four forward-pass requests and zero LLM tokens, with the annealing recipe costing approximately \$0.90 in GPU time. Extending this to Qwen3.5-2B, where layers and dose are selected using only the base model, the same recipe increases greedy GSM8K accuracy from 21.6\% to 69.4\%. Finally, as a proof of concept, we execute an eight-variable repair subproblem using QAOA on an IBM Heron processor.
\end{abstract}
\vspace{4pt}

%% file: figures/fig_teaser.tex
\begin{center}
\begin{tikzpicture}[x=1cm,y=1cm]
  \def\pw{5.25}\def\ph{3.35}\def\gap{0.30}
  \begin{scope}[shift={(0,0)}]
    \draw[panel=basegray] (0,0) rectangle (\pw,\ph);
    \node[panelhead=ink] at (0.22,\ph-0.18) {A\enspace The failure};
    \node[panelsub] at (0.22,\ph-0.52) {Qwen3.5-0.8B, thinking mode};
    \foreach \t/\x in {Let/0.55, {$x$}/1.10, {$=$}/1.62} {\node[tok=prodc] at (\x,1.62) {\t};}
    \foreach \t/\x in {so/2.30, {$x$}/2.82, {$=$}/3.34, {$2$}/3.86} {\node[tok=loopc] at (\x,1.62) {\t};}
    \node[font=\sffamily\tiny, text=softink] at (4.45,1.62) {\dots};
    \draw[arr, draw=loopc, rounded corners=6pt] (3.86,1.84) -- (3.86,2.14) -- (2.30,2.14) -- (2.30,1.84);
    \node[font=\sffamily\tiny, text=loopc] at (3.08,2.30) {repeats until the token cap};
    \node[stat] at (\pw/2,0.92) {93\% of MATH-500 answers reach the 32k cap};
    \node[stat, font=\sffamily\bfseries\scriptsize] at (\pw/2,0.45) {MATH-500 accuracy: 7.9\%};
  \end{scope}
  \begin{scope}[shift={(\pw+\gap,0)}]
    \draw[panel=agentorange] (0,0) rectangle (\pw,\ph);
    \node[panelhead=agentorange] at (0.22,\ph-0.18) {B\enspace LLM weight programmer};
    \node[panelsub] at (0.22,\ph-0.52) {GPT-6 Astra agent \citep{takehi2026sparks}};
    \node[card=agentorange, minimum width=1.75cm, minimum height=0.62cm] (ag) at (1.20,1.90) {reads weights,\\writes code};
    \node[card=basegray, minimum width=1.75cm, minimum height=0.62cm] (fp) at (4.05,1.90) {forward-pass\\requests};
    \draw[darr] (ag.east) -- node[above, font=\sffamily\tiny, text=softink] {50 calls} (fp.west);
    \node[font=\sffamily\tiny, text=ink, align=center] at (\pw/2,1.12) {dense rewrite of 64 columns in layers 15 and 19};
    \node[stat] at (\pw/2,0.62) {158 turns\enspace$\cdot$\enspace 89.7M tokens};
    \node[stat, font=\sffamily\bfseries\scriptsize, text=agentorange] at (\pw/2,0.27) {15.6 h\enspace$\cdot$\enspace \$312 per edit};
  \end{scope}
  \begin{scope}[shift={(2*\pw+2*\gap,0)}]
    \draw[panel=oursblue] (0,0) rectangle (\pw,\ph);
    \node[panelhead=oursblue] at (0.22,\ph-0.18) {C\enspace Tree-Ising (ours)};
    \node[panelsub] at (0.22,\ph-0.52) {optimization, forward passes only};
    \node[card=basegray, minimum width=1.25cm, minimum height=0.62cm] (st) at (0.90,1.90) {forward\\statistics};
    \begin{scope}[shift={(2.62,1.90)}]
      \foreach \i in {0,...,5} {\coordinate (q\i) at ({90+60*\i}:0.36);}
      \foreach \i in {0,...,5} {\foreach \j in {0,...,5} {\ifnum\i<\j \draw[oursblue!45, line width=0.3pt] (q\i) -- (q\j);\fi}}
      \foreach \i/\c in {0/oursblue,1/white,2/oursblue,3/white,4/white,5/oursblue} {\node[circle, draw=oursblue, fill=\c, line width=0.45pt, minimum size=0.15cm, inner sep=0pt] at (q\i) {};}
    \end{scope}
    \node[font=\sffamily\tiny, text=softink] at (2.62,1.30) {2,048 row QUBOs};
    \node[card=oursblue, minimum width=1.25cm, minimum height=0.62cm] (an) at (4.35,1.90) {GPU\\annealing};
    \draw[arr] (st.east) -- (2.14,1.90);
    \draw[arr] (3.10,1.90) -- (an.west);
    \node[font=\sffamily\tiny, text=ink, align=center] at (\pw/2,0.98) {sparse sign flips and exponent shifts;\\no shared-exponent or mantissa writes};
    \node[stat, font=\sffamily\bfseries\scriptsize, text=oursblue] at (\pw/2,0.40) {31 min\enspace$\cdot$\enspace \$0.90 per edit};
  \end{scope}
\end{tikzpicture}

\vspace{6pt}
\begin{tikzpicture}
\begin{groupplot}[group style={group size=4 by 1, horizontal sep=0.62cm}, minibar, width=4.55cm,
                  symbolic x coords={a,b,c}, enlarge x limits=0.42]
  \nextgroupplot[title={MATH-500@4\\accuracy (\%)}, ymax=68, ytick={0,20,40,60}, xtick={a,b,c},
                 xticklabels={Base,Agent,Ours}, /pgf/bar width=10pt, enlarge x limits=0.30]
    \addplot[fill=basegray, draw=basegray, bar shift=0pt] coordinates {(a,7.9)};
    \addplot[fill=agentorange, draw=agentorange, bar shift=0pt] coordinates {(b,51.7)};
    \addplot[fill=oursblue, draw=oursblue, bar shift=0pt] coordinates {(c,54.6)};
  \nextgroupplot[title={State-tree bits\\written (thousands)}, ymax=880, ytick={0,250,500,750}, xtick={a,b},
                 xticklabels={Agent,Ours}, nodes near coords={\pgfmathprintnumber[fixed,precision=0]{\pgfplotspointmeta}k}]
    \addplot[fill=agentorange, draw=agentorange, bar shift=0pt] coordinates {(a,720.6)};
    \addplot[fill=oursblue, draw=oursblue, bar shift=0pt] coordinates {(b,181.1)};
  \nextgroupplot[title={Cost of one run\\(USD, log scale)}, ymode=log, ymin=0.1, ymax=3000, log origin=infty,
                 ytick={0.1,1,10,100,1000}, yticklabels={0.1,1,10,100,1000}, xtick={a,b}, xticklabels={Agent,Ours},
                 nodes near coords={\$\pgfplotspointmeta},
                 point meta=explicit symbolic]
    \addplot[fill=agentorange, draw=agentorange, bar shift=0pt] coordinates {(a,312) [312]};
    \addplot[fill=oursblue, draw=oursblue, bar shift=0pt] coordinates {(b,0.90) [0.90]};
  \nextgroupplot[title={Qwen3.5-2B GSM8K\\greedy accuracy (\%)}, ymax=84, ytick={0,20,40,60,80}, xtick={a,b},
                 xticklabels={Base,Ours}]
    \addplot[fill=basegray, draw=basegray, bar shift=0pt] coordinates {(a,21.6)};
    \addplot[fill=oursblue, draw=oursblue, bar shift=0pt] coordinates {(b,69.4)};
\end{groupplot}
\end{tikzpicture}

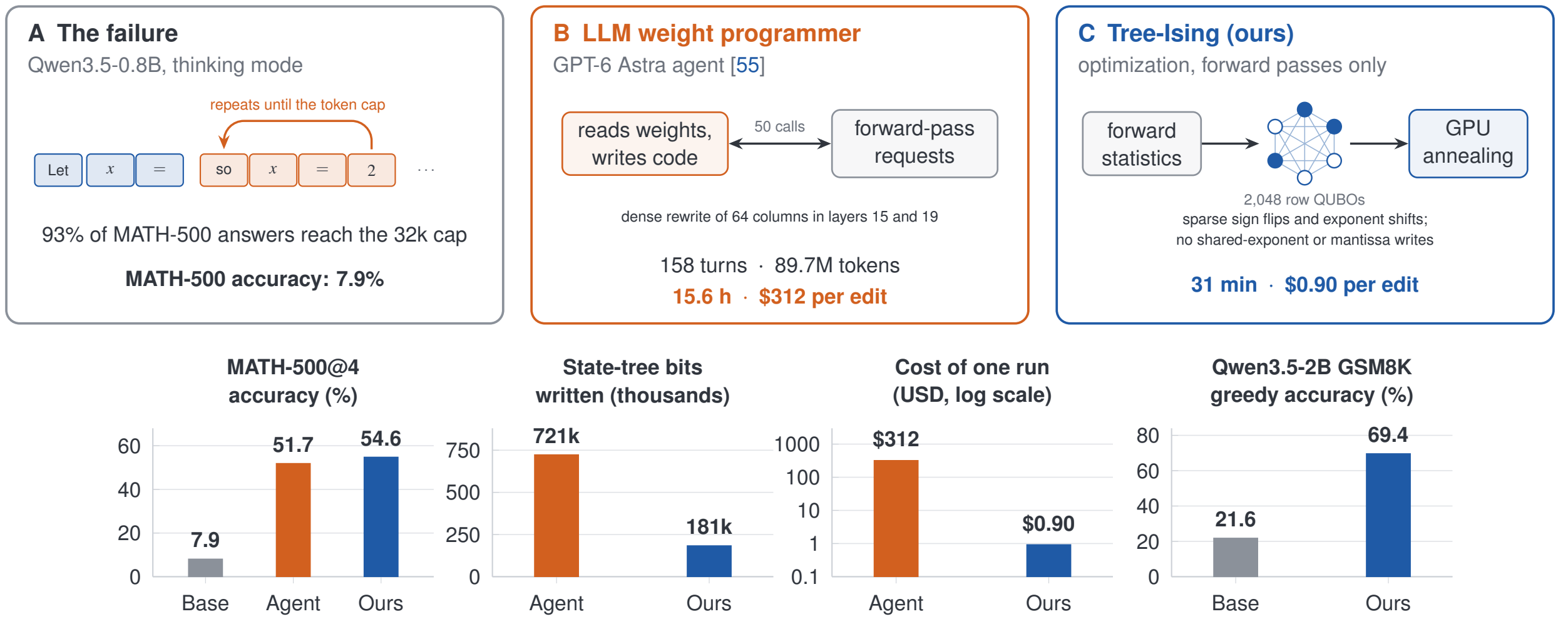
\captionof{figure}{\textbf{Two ways to repair a looping model.} (A) Qwen3.5-0.8B loops in thinking mode and rarely finishes. (B) An LLM agent
reads the weights, requests forward passes and writes a dense fix. (C) \ours finds the fix by optimization: one QUBO per
output row, solved by GPU annealing with sign flips and exponent shifts inside each group's state tree. Bottom: MATH-500
accuracy (base model as reported by \citet{takehi2026sparks}; agent and \ours under identical evaluation, \cref{tab:main}), bits written
(\cref{tab:footprint}), cost of the agent's discovery run and of one run of our fixed recipe (\cref{tab:cost}), and transfer without a reference edit (\cref{tab:transfer}).}
\label{fig:teaser}
\end{center}

%% file: sections/01_introduction.tex
\section{Introduction}
\label{sec:intro}
Weight-space programming has recently emerged as a compelling paradigm in model editing. \citet{takehi2026sparks} at MIT investigated whether a language model can program the weights of another network: reading its parameters, diagnosing a failure, and writing a fix directly into the weights, with an LLM agent acting as the programmer. Their most prominent demonstration targets a common failure mode in small reasoning models. In thinking mode, Qwen3.5-0.8B frequently falls into a short cycle of repeated text, continuing until the token budget is exhausted; this occurs on 93\% of MATH-500 problems under a 32,768-token cap, limiting its accuracy to merely 7.9\%. A GPT-6 Astra agent, provided with the checkpoint, practice problems, and a budget of 50 forward-pass requests, traced these loops to the MLPs of layers 15 and 19 and rewrote 64 columns in each of their output projections. Consequently, MATH-500 accuracy rose to 48.1\% without a single gradient step.
While this result proves that weights can be programmed, it also highlights the substantial price of doing so with an agent: 158 turns, 89.7 million tokens from a frontier model, 50 forward-pass requests, and 15.6 hours of compute for a single edit (\cref{tab:cost}). Crucially, every step of this process relies on judgment calls made in natural language, offering no guarantee that the resulting edit is the optimal one available.
In this work, we cast the repair as a sparse discrete optimization problem over the stored weights. The evidence gathered by the agent---forward passes on looping answers versus successful ones---can be formalized as a mathematical objective, where an edit is defined as a discrete choice among changes to stored bits. We build this formulation on state trees~\citep{ali2026everyweight}, a general representation of neural networks in which every weight is a path of discrete decisions through a tree, and weights sharing the same decision share the recording node. Lossless BF16 storage is a special case of this representation, where each group of 64 weights shares its exponent as a node; this is the specific case we utilize. Our formulation, which we call \ours, comprises five key steps (\cref{fig:method}):
\begin{enumerate}[label=\arabic*.]
\item \textbf{Represent the weights as state trees.} In the lossless BF16 state tree, multiplying a weight by a power of two or flipping its sign alters only its 4-bit state and sign bit. Moves that preserve each group's maximum leave the shared exponent untouched. Preselecting one such move per candidate weight reduces the edit to a series of binary decisions.
\item \textbf{Select the layers.} A single forward scan of the base model evaluates how strongly each MLP writes along the loop direction, automatically identifying the layers to edit without requiring an agent.
\item \textbf{Define the objective.} Forward statistics yield a loop direction and a concluding direction. The objective is to minimize the layer's projection along their combination on loop tokens, preserve the output on productive tokens, and restrict the total number of modified weights.
\item \textbf{Formulate as an Ising model.} For fixed statistics and preselected moves, the objective decomposes exactly into independent subproblems, one per output row. Each row subproblem is formulated as a dense QUBO, equivalently an Ising model.
\item \textbf{Solve the optimization.} A GPU solves all row problems for a given layer simultaneously using simulated annealing or simulated bifurcation, a quantum-inspired Ising solver. Because these are native Ising models, they also map directly onto quantum annealers and QAOA.
\end{enumerate}
\input{tables/tab_glance}
Evaluated on the two layers targeted by the agent, \ours achieves 54.6\% on MATH-500, outperforming the agent's edit (51.7\%) under an identical evaluation protocol, while the base model scores only 7.9\% in the evaluation of \citet{takehi2026sparks}. Notably, it outperforms the agent edit across all five math benchmarks tested, with statistically significant gains on MATH-500 and GSM8K after correcting for multiple comparisons (\cref{tab:glance,tab:main}). Furthermore, replacing the annealer with simulated bifurcation~\citep{goto2019sb,goto2021sb} yields an equally effective repair, achieving 55.1\% on MATH-500 with 4\% fewer modified weights, demonstrating that the results do not hinge on a single solver. Crucially, this approach requires only 4 forward-pass requests instead of 50, reducing forward-pass compute by a factor of 34 and eliminating the need for LLM tokens entirely. The entire recipe executes in 31 minutes at a cost of approximately \$0.90. Moreover, the resulting edit is substantially smaller: it modifies 13\% fewer weights, reduces BF16 bit flips by a factor of 2.2, cuts state-tree bit changes by a factor of 4, and, by construction of its move set, leaves all shared exponents and mantissas untouched.
It is important to clarify the scope of this comparison. The agent incurred a significant cost to discover the failure, identify the direction, and select the layers. In contrast, our recipe was designed after analyzing its public logs; on Qwen3.5-0.8B, it reuses the agent's two selected layers and calibrates its dose based on the agent's edit. Thus, the cost figures compare the exhaustive process of finding a fix against the application of a fixed, optimized recipe. Our primary aim is to achieve a superior repair through fewer weight modifications. For context, a dense closed-form solution of the same objective, without sparsity or discrete moves, reaches similar accuracy on our practice screen but modifies 96\% of the weights in both matrices---62 times as many as \ours (\cref{sec:exp-ablations}).
Extending this recipe to Qwen3.5-2B demonstrates that the agent is not strictly necessary. With layers selected by the forward scan and the dose calibrated from the base model alone, the method raises greedy GSM8K accuracy from 21.6\% to 69.4\%.
\paragraph{Contributions.}
The contributions of this work are summarized as follows:
\begin{itemize}
\item We propose a sparse discrete formulation for model repair on state trees, introducing tree-local moves, a forward-only objective, and an exact decomposition into independent row Ising problems for fixed statistics and preselected moves (\cref{sec:method}).
\item We demonstrate that on Qwen3.5-0.8B, \ours produces an edit that is more accurate than the agent-based repair of \citet{takehi2026sparks}, while modifying fewer weights and bits, at a small fraction of the forward passes, time, and computational cost (\cref{sec:exp}).
\item We show that this recipe transfers effectively to Qwen3.5-2B without any reference edit, achieving a 47.8-point gain in GSM8K accuracy under greedy decoding (\cref{sec:transfer}).
\end{itemize}
Finally, as a supplementary proof of concept, we successfully execute an eight-variable repair subproblem using QAOA on an IBM Heron quantum processor (\cref{sec:quantum}).

%% file: tables/tab_glance.tex
\begin{table}[t]
\centering
\caption{\textbf{\ours against the LLM agent on Qwen3.5-0.8B, at a glance.} Same model, same two layers, same evaluation harness; accuracy uses four completions per MATH-500 problem. Cost
compares the agent's discovery run with one application of our fixed recipe, which reuses the agent's layers and dose
(\cref{sec:exp-cost}). Zero shared-exponent and mantissa writes hold by construction of the move set. \ours is solved by annealing here;
simulated bifurcation gives an equivalent edit (55.1\% on MATH-500, 108,327 changed weights). Details in
\cref{tab:main,tab:cost,tab:footprint}.}
\label{tab:glance}
\small
\begin{tabular}{@{}llrrl@{}}
\toprule
 & & Agent edit~\citep{takehi2026sparks} & \ours (ours) & \ours advantage \\
\midrule
\multirow{3}{*}{Quality}
  & MATH-500 accuracy (\%)              & 51.7          & \textbf{54.6}   & $+2.9$ points, $p\approx0.003$ \\
  & Benchmarks ahead (of 5)             & --            & \textbf{5}      & 2 significant after correction \\
  & MATH-500 answers cut at cap (\%)    & 31.6          & \textbf{23.9}   & 1.3$\times$ fewer \\
\midrule
\multirow{5}{*}{Cost}
  & Forward-pass requests               & 50            & \textbf{4}      & 12.5$\times$ fewer \\
  & Forward-pass GPU time               & 12.2 h        & \textbf{21.8 min} & 34$\times$ less \\
  & LLM tokens                          & 89.7M         & \textbf{0}      & none needed \\
  & Wall clock                          & 15.6 h        & \textbf{31 min} & 30$\times$ faster \\
  & Cost                                & \$312         & \textbf{\$0.90} & about 350$\times$ cheaper \\
\midrule
\multirow{5}{*}{Edit}
  & Changed weights                     & 130,188       & \textbf{112,630} & 13\% fewer \\
  & BF16 bit flips                      & 621,388       & \textbf{278,767} & 2.2$\times$ fewer \\
  & State-tree bit flips                & 720,586       & \textbf{181,062} & 4.0$\times$ fewer \\
  & Shared-exponent writes              & 1,261         & \textbf{0}       & none (by design) \\
  & Mantissa writes                     & 129,668       & \textbf{0}       & none (by design) \\
\bottomrule
\end{tabular}
\end{table}

%% file: sections/02_background.tex
\section{Background}
\label{sec:background}
\paragraph{Reasoning loops.}
We define an answer as \emph{cut} when the generation process reaches the maximum token budget before the model produces its final conclusion. In the models evaluated in this work, nearly all cut answers exhibit looping behavior: within the final 2,000 tokens, the majority of 8-grams have already appeared earlier in the same sequence. While repetition in neural text generation has been extensively studied through decoding heuristics and modified training objectives~\citep{holtzman2020curious,welleck2020unlikelihood,xu2022loop}, and recent reports on reasoning models analyze how extended thinking traces interact with decoding configurations~\citep{deepseek2025r1,qwen35card}, we frame looping as a property of the network's weights. This perspective treats the failure mode as an artifact that a direct weight edit can resolve.
\paragraph{Weight-space programming.}
\citet{takehi2026sparks} enabled a frontier LLM agent to read the parameters of Qwen3.5-0.8B and execute forward passes on practice problems. 
Within a budget of 50 requests, the agent identified a specific direction in the residual stream that remains active during loops, traced this behavior to the MLP output projections (\code{down\_proj}) of layers 15 and 19,
and applied a structured modification to 64 columns in each matrix. The comprehensive logs released by the authors~\citep{takehi2026logs} detail every interaction, token count, and forward pass, allowing us to precisely quantify the computational cost of the procedure (\cref{sec:exp-cost}). We denote the resulting checkpoint as the \emph{agent edit}. Crucially, the agent's workflow encompasses three distinct phases, which our formulation renders mathematically explicit: identifying the target layers, defining the optimization objective, and searching for the precise weight modifications required to achieve it.

%% file: sections/03_method.tex
\section{Method: the repair as a sparse Ising problem}
\label{sec:method}

\ours edits frozen MLP output projections $W\in\mathbb{R}^{d\times n}$, where $d$ denotes the number of output rows and $n$ the number of input neurons. We cast this edit as a discrete optimization problem comprising five key steps (\cref{fig:method}): representing the weights and admissible moves (\cref{sec:step1}), selecting the target layers (\cref{sec:step2}), defining the optimization objective (\cref{sec:step3}), formulating the problem as independent Ising models (\cref{sec:step4}), and solving them (\cref{sec:step5}).

\input{figures/fig_method}

\subsection{Step 1: a QUBO-friendly representation of the weights}
\label{sec:step1}

\paragraph{State trees.} State trees~\citep{ali2026everyweight} provide a general representation for neural networks. Every weight is defined as a path of discrete decisions from the root of a tree to a leaf, and weights that share the same decision share the node that records it. This representation seamlessly unifies low-bit compression and exact storage. Lossless BF16 storage is a specific instance of this framework, and it is the one we employ. A BF16 number consists of one sign bit, an 8-bit exponent $e$, and a 7-bit mantissa~\citep{kalamkar2019bf16}. The lossless BF16 state tree groups 64 consecutive weights within a row and stores their maximum exponent $E$ once as a shared node. Each individual weight then stores a 4-bit state $D=E-e\in\{0,\dots,14\}$ and a leaf containing its sign and mantissa; state 15 designates a rare escape whose exponent is maintained in a side list. This instantiation reproduces every weight exactly, cleanly separating what a group shares (its exponent) from what each weight owns (its offset, sign, and mantissa), mirroring the structure of block floating-point and microscaling formats in their lossy counterparts~\citep{drumond2018hbfp,rouhani2023mx}. Because every stored quantity is a discrete decision, an edit is fundamentally a change of decisions.

\paragraph{Tree-local moves.} We restrict each weight to be multiplied by a factor from the set
\begin{equation}
  \Fset=\bigl\{-1,\;2,\;\tfrac12,\;4,\;-2,\;-4,\;-\tfrac12,\;\tfrac14\bigr\}
  \label{eq:moves}
\end{equation}
or to remain unchanged. For a standard weight with state $D$, a factor with $k=\log_2|f|$ shifts the state to $D-k$ and flips the sign if $f<0$. A move is deemed admissible only if it preserves the group's shared exponent and generates no escape. Growth ($k>0$) requires $D\ge k$, ensuring no weight exceeds the group maximum; shrinkage ($k<0$) requires $D\ge1$ and $D-k\le14$, guaranteeing that the group maximum is never reduced and no weight falls outside the 4-bit range. Sign flips are universally admissible. Every admissible move is exact in BF16, preserves the mantissa, and alters at most the sign bit and the 4-bit state of a single weight. For instance, the weight $0.125$ in a group with $E=127$ has $D=3$; doubling it yields $0.25$ with $D=2$, resulting in a change to a single stored bit.

\paragraph{Binary formulation and the role of the state tree.} The binary nature of the formulation arises from preselecting one move per candidate weight (\cref{sec:step4}): the sole remaining decision is whether to apply it, rendering the edit a vector of binary variables where every subsequent quantity is a polynomial of degree at most two, conforming to the QUBO format. Multiplying by a power of two is also exact in standard BF16, meaning the tree itself does not create the binary structure. Rather, its contribution lies in providing a format-native move set and its corresponding accounting. Because admissible moves never alter the group maximum, the shared node is never rewritten, and the number of stored bits modified by an edit can be directly read from the tree. The guarantee that the edit writes no shared exponent and no mantissa therefore holds by construction, rather than as an empirical observation.

\subsection{Step 2: selecting the layers}
\label{sec:step2}

\paragraph{Forward statistics.} The base model generates an answer for each practice problem once, subject to an 8,192-token cap. Reference answers denote correct, completed traces, while a trace is classified as looping if it reaches the cap and more than 80\% of the 8-grams in its tail repeat. Teacher-forced passes over these texts record, for every layer, the residual stream $h(t)$ after the layer, the output $m(t)$ of its MLP, and the input $a(t)$ of its output projection, across the four token sets defined in \cref{tab:tokensets}.

\begin{table}[h]
\centering
\caption{Token sets used by the layer scan and the objective. All sets come from the base model's own answers to practice problems.}
\label{tab:tokensets}
\small
\begin{tabularx}{\textwidth}{@{}l l Y@{}}
\toprule
Set & Role & Definition \\
\midrule
$\R$     & loop      & In a looping trace, find the first 256-token window in which at least 80\% of 8-grams repeat, after at least
                       300 generated tokens. In the next 1,000 positions, take the final token of every repeated 8-gram. \\
$\Pset$  & protected & The 2,000 to 50 tokens before each loop onset, and every generated token of correct completed traces. \\
$\Zset$  & concluding & The last 64 thinking tokens before \code{</think>} in a correct trace with at least 200 thinking tokens. \\
$\Mset$  & continuing & The earlier thinking tokens of the same correct traces. \\
\bottomrule
\end{tabularx}
\end{table}

The mean residuals after layer $\ell$ yield two unit directions and their combination,
\begin{equation}
  u_{\ell,\mathrm{loop}}\propto\bar h_{\ell,\R}-\bar h_{\ell,\Pset},\qquad
  u_{\ell,\mathrm{end}}\propto\bar h_{\ell,\Zset}-\bar h_{\ell,\Mset},\qquad
  u_\ell\propto u_{\ell,\mathrm{loop}}-\beta\,u_{\ell,\mathrm{end}},
  \label{eq:directions}
\end{equation}
with $\beta=0.5$. Reducing the projection of a layer's output onto $u_\ell$ weakens the loop and encourages the model to conclude.

\paragraph{Layer scan.} The write score of layer $\ell$ quantifies how much more its MLP pushes the residual stream along $u_\ell$ on loop tokens compared to productive ones,
\begin{equation}
  s_\ell=u_\ell^{\top}\bigl(\bar m_{\ell,\R}-\bar m_{\ell,\Pset}\bigr).
  \label{eq:scan}
\end{equation}
Our selection rule targets the two highest-scoring layers, excluding the final layer, whose MLP writes directly into the unembedding. On Qwen3.5-0.8B, three layers stand out significantly (\cref{fig:layerscan}, top): layer 23 (the final layer), and layers 19 and 15, each scoring at least twice as high as any other layer. Consequently, the rule selects layers 15 and 19, exactly matching the agent's choice. We established this exclusion rule after observing these scores and subsequently applied it unchanged to Qwen3.5-2B, where it again selects layers 15 and 19 (\cref{fig:layerscan}, bottom). In contrast, a gradient-based attribution, which scores layers by the first-order effect of removing their MLP output on the log-probability of continuing a loop, ranks layers 14, 21, 18, and 16 highest, failing to recover the agent's choice.

Both models utilize a hybrid layout where every fourth block (layers 3, 7, 11, 15, 19, and 23) employs full attention, while the others use linear attention. In both models, the three highest-scoring layers are full-attention blocks (shaded in \cref{fig:layerscan}). The agreement between the two models may therefore reflect their shared architectural design rather than providing independent evidence that the scan identifies the optimal layers. Nevertheless, this alignment is suggestive in itself, since only full attention can copy exactly from distant positions in the context, which is precisely what a loop requires. Evaluating the scan on a model without this specific layout remains a direction for future work.

\input{figures/fig_layerscan}

\subsection{Step 3: the objective}
\label{sec:step3}

For a given edited layer, we drop the layer index $\ell$ and denote its target direction as $u$. The objective requires the mean input on loop tokens, $\mu_{\R}=\E_{t\in\R}[a(t)]$, and the second moment of inputs on protected tokens, $C_{\Pset}=\E_{t\in\Pset}[a(t)a(t)^{\top}]$. For a fixed input $a(t)$, an edit $\Delta$ alters the layer output by $\Delta a(t)$. We quantify the effect of the edit along $u$ on loop tokens, termed its \emph{dose}, and its perturbation on protected tokens, termed its \emph{collateral}:
\begin{equation}
  \dose(\Delta)=u^{\top}\Delta\,\mu_{\R},\qquad
  \coll(\Delta)=\E_{t\in\Pset}\bigl\|\Delta a(t)\bigr\|^2=\sum_{r=1}^{d}\Delta_{r:}\,C_{\Pset}\,\Delta_{r:}^{\top}.
  \label{eq:dose-coll}
\end{equation}
This collateral term corresponds to the layer-output reconstruction error commonly employed in pruning and quantization~\citep{hassibi1993obs,frantar2023gptq}. For scaling factors $\gamma>0$ and $\lambda\ge0$, the edit minimizes
\begin{equation}
  \mathcal{E}(\Delta)=\gamma\,\dose(\Delta)+\coll(\Delta)+\lambda\,\|\Delta\|_0
  \qquad\text{over edits constructed from admissible moves.}
  \label{eq:energy}
\end{equation}
The parameter $\gamma$ is calibrated to ensure the edit achieves a target dose, while $\lambda$ restricts the edit to a specific budget of changed weights (\cref{sec:step5}). The target is defined relative to the layer's own write along $u$ on loop tokens, $d_{\mathrm{target}}=-\tau\,\bigl|u^{\top}W\mu_{\R}\bigr|$. On Qwen3.5-0.8B, we set the target to 1.6 times the dose of the agent's edit along the same $u$, a multiplier fixed during development prior to the dose sweep in \cref{tab:ablations}; this corresponds to $\tau=6.7$ for layer 15 and $\tau=6.1$ for layer 19 (the achieved doses are 6.8 and 6.2 times the own write, respectively, as shown in \cref{tab:dose}). The comparison with the agent is therefore not strictly dose-matched. For a novel model, we use $\tau=6.5$, ensuring the entire objective is derived from the base model alone. Crucially, the formulation relies exclusively on forward passes, requiring no gradients or training.

\subsection{Step 4: exact decomposition into row Ising models}
\label{sec:step4}

Each of the three terms in \cref{eq:energy} decomposes into sums over the output rows: the dose is linear in $\Delta$, the collateral is a sum of row-wise quadratic forms, and the sparsity penalty counts nonzeros row by row. For fixed statistics and penalties, the problem therefore separates exactly into one independent problem per row. The weight budget couples the rows only through their aggregate total, which is managed by the outer adjustment of $\lambda$.

Within each layer, we retain the $K$ columns exhibiting the largest scores $(u^{\top}W_{:j})\,\mu_{\R,j}$, corresponding to the neurons whose current write along $u$ on loop tokens is most significant. Because this score is independent of the row, every row of a given layer utilizes the same $K$ columns. For each retained weight $(r,j_a)$, we preselect the admissible factor $f_a$ that yields the lowest single-weight energy at the current $\gamma$ and $\lambda$. We define $\delta_a=(f_a-1)W_{rj_a}$ and let $x_a\in\{0,1\}$ act as a switch for the move. The row problem is then formulated as the QUBO
\begin{equation}
  \mathcal{E}_r(x)=\sum_{a=1}^{K}h_a x_a+\sum_{a\neq b}J_{ab}x_a x_b,\qquad
  h_a=\gamma\,u_r\,\mu_{\R,j_a}\,\delta_a+C_{\Pset,j_aj_a}\,\delta_a^2+\lambda,\qquad
  J_{ab}=\delta_a\,\delta_b\,C_{\Pset,j_aj_b}.
  \label{eq:qubo}
\end{equation}
The linear field $h_a$ aggregates the dose gained, the self-collateral, and the sparsity cost of a single move; the coupling $J_{ab}$ represents the collateral generated jointly by two moves. By substituting spins $s=2x-1$, the same energy is expressed as an Ising model,
\begin{equation}
  \mathcal{E}_r(s)=\mathrm{const}+\sum_a H_a s_a+\sum_{a<b}I_{ab}s_as_b,\qquad
  H_a=\tfrac12 h_a+\tfrac12\textstyle\sum_{b\neq a}J_{ab},\qquad I_{ab}=\tfrac12 J_{ab},
  \label{eq:ising}
\end{equation}
where $J$ is symmetric and enters \cref{eq:qubo} through the ordered-pair sum. This decomposition is exact given the retained columns and the preselected moves; both are heuristics designed to shrink the search space, and we repeat the preselection whenever $\gamma$ is updated. With $K=512$, each edited layer of Qwen3.5-0.8B yields 1,024 dense Ising problems with 512 spins each, resulting in 2,048 problems across the two edited layers.

\subsection{Step 5: solving}
\label{sec:step5}

\paragraph{GPU annealing.} Our default solver processes all rows of a given layer as a single GPU batch via simulated annealing~\citep{kirkpatrick1983sa}. The final solve executes 16 replicas per row, starting from the zero edit, for 300 sweeps of single-spin Metropolis updates in random order. The temperature decreases geometrically from 2 to $10^{-4}$ times the row's median $|h_a|$, and we retain the lowest-energy replica.

\paragraph{Simulated bifurcation.} As an alternative solver, we employ discrete simulated bifurcation (dSB)~\citep{goto2019sb,goto2021sb}, a quantum-inspired method that integrates the classical dynamics of coupled bifurcating oscillators and reads spins from their signs. Each replica initializes near the zero edit; the pump ramps up over the first quarter of 1,200 steps of size 0.5, and the best sampled state of every trajectory, refined by single-flip descent in float64, is retained. Rows are processed in batches of 32, maintaining the same replica counts as in annealing. All other components, including the candidates, the preselected moves, and the calibration protocol, remain unchanged, ensuring the two solvers differ solely in how each row problem is searched.

\paragraph{Calibration.} Two global scalars are calibrated prior to solving the individual row problems. The sparsity weight initializes at $\lambda_0=2\operatorname{median}C_{\Pset,jj}\,\delta^2$, representing the median single-weight collateral over the retained weights and the three basic moves $f\in\{-1,2,\tfrac12\}$; the candidate moves themselves always utilize all eight factors. At each value of $\lambda$, a geometric bisection over $\gamma\in[10^{-6},10^{6}]$, utilizing quick solves of 4 replicas and 60 sweeps, identifies the smallest $\gamma$ whose edit reaches the target dose, and the final solve adopts that $\gamma$. If the layer subsequently exceeds its budget, $\lambda$ is doubled and the $\gamma$ search is repeated for at most six attempts, ensuring the dose target is re-met after every adjustment to $\lambda$. The statistics remain fixed throughout this process. \Cref{tab:dose} reports the achieved dose and edit count for every edited layer.

\paragraph{Alternative solvers.} Crucially, the formulation in Steps 1 through 4 is entirely solver-agnostic. The row problems are standard Ising models and can be seamlessly handed to any Ising solver, ranging from greedy descent to quantum annealing~\citep{kadowaki1998qa,johnson2011qa} or QAOA~\citep{farhi2014qaoa}. We examine the quantum route in \cref{sec:quantum}.

%% file: figures/fig_method.tex
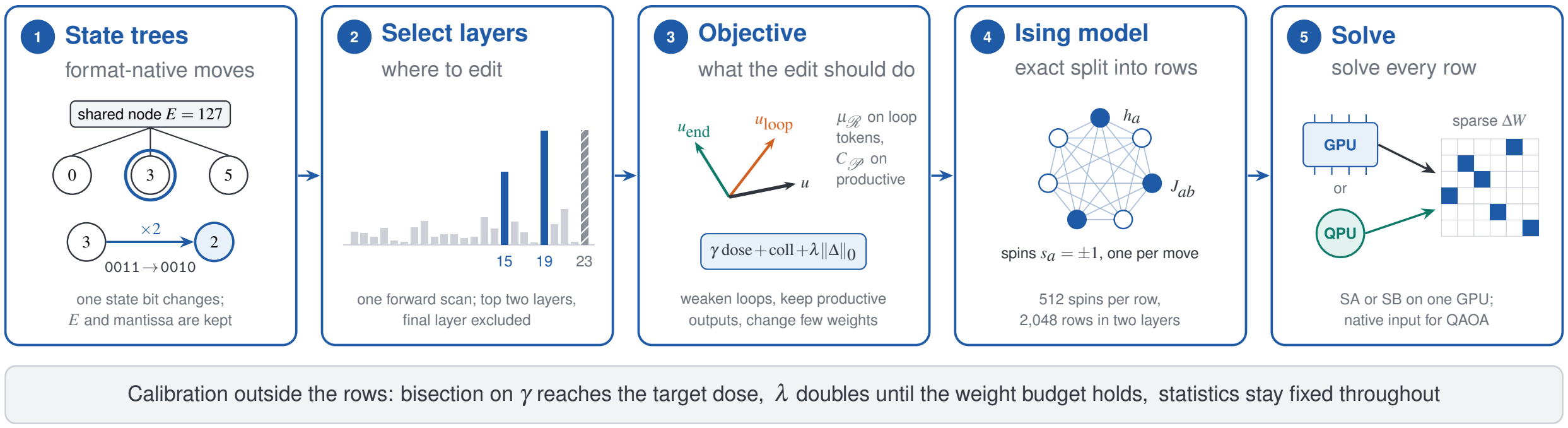
\begin{figure}[t]
\centering
\begin{tikzpicture}[x=1cm,y=1cm]
  \def\pw{3.05}\def\ph{3.55}\def\gap{0.27}
  \foreach \i/\t/\s in {1/{State trees}/{format-native moves},
                        2/{Select layers}/{where to edit},
                        3/{Objective}/{what the edit should do},
                        4/{Ising model}/{exact split into rows},
                        5/{Solve}/{solve every row}} {
    \begin{scope}[shift={({(\i-1)*(\pw+\gap)},0)}]
      \draw[panel=oursblue] (0,0) rectangle (\pw,\ph);
      \node[circle, fill=oursblue, text=white, font=\sffamily\bfseries\tiny, minimum size=0.36cm, inner sep=0pt]
        at (0.34,\ph-0.32) {\i};
      \node[font=\sffamily\bfseries\footnotesize, text=oursblue, anchor=west, inner sep=0pt] at (0.62,\ph-0.30) {\t};
      \node[panelsub, anchor=west] at (0.62,\ph-0.66) {\s};
    \end{scope}
  }
  \foreach \i in {1,...,4} {
    \draw[arr, draw=oursblue] ({\i*\pw+(\i-1)*\gap+0.02},\ph/2) -- ({\i*(\pw+\gap)-0.02},\ph/2);
  }
  \begin{scope}[shift={(0,0)}]
    \node[draw=ink, fill=basefill, rounded corners=1.5pt, font=\sffamily\tiny, inner sep=2.2pt] (E) at (1.52,2.42)
      {shared node $E=127$};
    \foreach \d/\x/\n in {0/0.70/c0, 3/1.52/c1, 5/2.34/c2} {
      \node[circle, draw=ink, fill=white, line width=0.45pt, minimum size=0.40cm, inner sep=0pt, font=\sffamily\tiny]
        (\n) at (\x,1.78) {$\d$};
      \draw[ink, line width=0.4pt] (E.south) -- (\n.north);
    }
    \draw[oursblue, line width=0.9pt] (c1) circle (0.26);
    \node[circle, draw=ink, fill=white, line width=0.45pt, minimum size=0.40cm, inner sep=0pt, font=\sffamily\tiny]
      (b3) at (0.85,1.08) {$3$};
    \node[circle, draw=oursblue, fill=oursfill, line width=0.7pt, minimum size=0.40cm, inner sep=0pt, font=\sffamily\tiny]
      (a2) at (2.19,1.08) {$2$};
    \draw[arr, draw=oursblue] (b3.east) -- node[above, font=\sffamily\tiny, text=oursblue, inner sep=1.5pt] {$\times2$} (a2.west);
    \node[font=\sffamily\tiny, text=ink] at (1.52,0.82) {\code{0011}$\,\to\,$\code{0010}};
    \node[font=\sffamily\tiny, text=softink, align=center] at (1.52,0.36) {one state bit changes;\\
      $E$ and mantissa are kept};
  \end{scope}
  \begin{scope}[shift={(\pw+\gap,0)}]
    \draw[rule, line width=0.4pt] (0.22,1.05) -- (2.86,1.05);
    \foreach \l/\s in {0/0.0384,1/0.0342,2/0.0221,3/0.0488,4/0.0116,5/0.0074,6/0.0512,7/0.0691,8/0.0233,9/0.0300,
                       10/0.0295,11/0.0189,12/0.0234,13/0.0802,14/0.0474,16/0.0675,17/0.0352,18/0.0,20/0.0588,
                       21/0.1043,22/0.0089} {
      \fill[rule] ({0.30+0.105*\l},1.05) rectangle ++(0.075,{max(3.6*\s,0.02)});
    }
    \foreach \l/\s in {15/0.2131,19/0.3319} {
      \fill[oursblue] ({0.30+0.105*\l},1.05) rectangle ++(0.075,{3.6*\s});
      \node[font=\sffamily\tiny, text=oursblue] at ({0.3375+0.105*\l},0.88) {\l};
    }
    \fill[basegray, postaction={pattern=north east lines, pattern color=white}]
      ({0.30+0.105*23},1.05) rectangle ++(0.075,{3.6*0.3337});
    \node[font=\sffamily\tiny, text=softink] at ({0.3375+0.105*23},0.88) {23};
    \node[font=\sffamily\tiny, text=softink, align=center] at (1.52,0.36) {one forward scan; top two layers,\\
      final layer excluded};
  \end{scope}
  \begin{scope}[shift={(2*\pw+2*\gap,0)}]
    \begin{scope}[shift={(0.95,1.55)}]
      \draw[-{Stealth[length=3.5pt]}, line width=0.8pt, draw=loopc] (0,0) -- (0.48,0.62)
        node[above, font=\sffamily\tiny, text=loopc, inner sep=1pt] {$\uloop$};
      \draw[-{Stealth[length=3.5pt]}, line width=0.8pt, draw=endc] (0,0) -- (-0.36,0.58)
        node[above, font=\sffamily\tiny, text=endc, inner sep=1pt] {$\uend$};
      \draw[-{Stealth[length=3.5pt]}, line width=1.0pt, draw=ink] (0,0) -- (0.70,0.14)
        node[right, font=\sffamily\tiny, text=ink, inner sep=1pt] {$u$};
    \end{scope}
    \node[font=\sffamily\tiny, text=softink, align=left, anchor=west] at (1.95,2.05) {$\mu_{\R}$ on loop\\tokens,\\$C_{\Pset}$ on\\productive};
    \node[draw=oursblue, fill=oursfill, rounded corners=2pt, font=\sffamily\tiny, text=ink, inner sep=2.6pt]
      at (1.52,0.98) {$\gamma\,\dose+\coll+\lambda\|\Delta\|_0$};
    \node[font=\sffamily\tiny, text=softink, align=center] at (1.52,0.36) {weaken loops, keep productive\\
      outputs, change few weights};
  \end{scope}
  \begin{scope}[shift={(3*\pw+3*\gap,0)}]
    \begin{scope}[shift={(1.52,1.82)}]
      \foreach \i in {0,...,6} {\coordinate (r\i) at ({90+360/7*\i}:0.56);}
      \foreach \i in {0,...,6} {\foreach \j in {0,...,6} {\ifnum\i<\j \draw[oursblue!40, line width=0.3pt] (r\i) -- (r\j);\fi}}
      \foreach \i/\c in {0/oursblue,1/white,2/white,3/oursblue,4/white,5/oursblue,6/white} {
        \node[circle, draw=oursblue, fill=\c, line width=0.45pt, minimum size=0.19cm, inner sep=0pt] at (r\i) {};}
      \node[font=\sffamily\tiny, text=ink, anchor=west] at ($(r0)+(0.12,0.02)$) {$h_a$};
      \node[font=\sffamily\tiny, text=ink, anchor=west] at (0.62,-0.18) {$J_{ab}$};
    \end{scope}
    \node[font=\sffamily\tiny, text=ink] at (1.52,0.95) {spins $s_a=\pm1$, one per move};
    \node[font=\sffamily\tiny, text=softink, align=center] at (1.52,0.36) {512 spins per row,\\2,048 rows in two layers};
  \end{scope}
  \begin{scope}[shift={(4*\pw+4*\gap,0)}]
    \node[draw=oursblue, fill=oursfill, rounded corners=1.5pt, minimum width=0.78cm, minimum height=0.46cm,
          font=\sffamily\bfseries\tiny, text=oursblue, inner sep=0pt] (gpu) at (0.72,2.10) {GPU};
    \foreach \dx in {-0.27,-0.09,0.09,0.27} {
      \draw[oursblue, line width=0.4pt] ($(gpu.north)+(\dx,0)$) -- ++(0,0.08);
      \draw[oursblue, line width=0.4pt] ($(gpu.south)+(\dx,0)$) -- ++(0,-0.08);
    }
    \node[font=\sffamily\tiny, text=softink] at (0.72,1.63) {or};
    \node[circle, draw=qteal, fill=qfill, line width=0.6pt, minimum size=0.50cm, inner sep=0pt,
          font=\sffamily\bfseries\tiny, text=qteal] (qpu) at (0.72,1.17) {QPU};
    \begin{scope}[shift={(1.78,1.14)}]
      \foreach \i in {0,...,5} {\foreach \j in {0,...,5} {
        \draw[rule, fill=white, line width=0.2pt] (\i*0.17,\j*0.17) rectangle ++(0.17,0.17);}}
      \foreach \i/\j in {1/4, 3/1, 4/5, 0/2, 5/0, 2/3} {\fill[oursblue] (\i*0.17,\j*0.17) rectangle ++(0.17,0.17);}
      \node[font=\sffamily\tiny, text=softink] at (0.51,1.20) {sparse $\Delta W$};
    \end{scope}
    \draw[arr] (gpu.east) -- (1.72,1.80);
    \draw[arr, draw=qteal] (qpu.east) -- (1.72,1.42);
    \node[font=\sffamily\tiny, text=softink, align=center] at (1.52,0.36) {SA or SB on one GPU;\\ native input for QAOA};
  \end{scope}
  \draw[rounded corners=3pt, draw=rule, fill=basefill, line width=0.5pt] (0,-0.82) rectangle (5*\pw+4*\gap,-0.22);
  \node[font=\sffamily\scriptsize, text=ink] at ({(5*\pw+4*\gap)/2},-0.52) {Calibration outside the rows:
    bisection on $\gamma$ reaches the target dose,\enspace $\lambda$ doubles until the weight budget holds,\enspace
    statistics stay fixed throughout};
\end{tikzpicture}
\caption{\textbf{\ours in five steps.} (1)~In the lossless BF16 state tree, admissible moves keep the group's shared exponent and the mantissa; multiplying a
weight by two changes one stored state bit. One move is preselected per candidate weight, so the edit becomes binary. (2)~One forward scan of the base
model scores each MLP's write along the loop direction and selects layers 15 and 19 (\cref{fig:layerscan}). (3)~Forward
statistics define the dose along $u$, the collateral change on productive tokens and a sparsity cost. (4)~For fixed statistics
the objective splits exactly into one dense Ising model per output row. (5)~GPU simulated annealing or simulated bifurcation solves all rows of a layer at once; the same models are the native
input of quantum annealers and QAOA.}
\label{fig:method}
\end{figure}

%% file: figures/fig_layerscan.tex
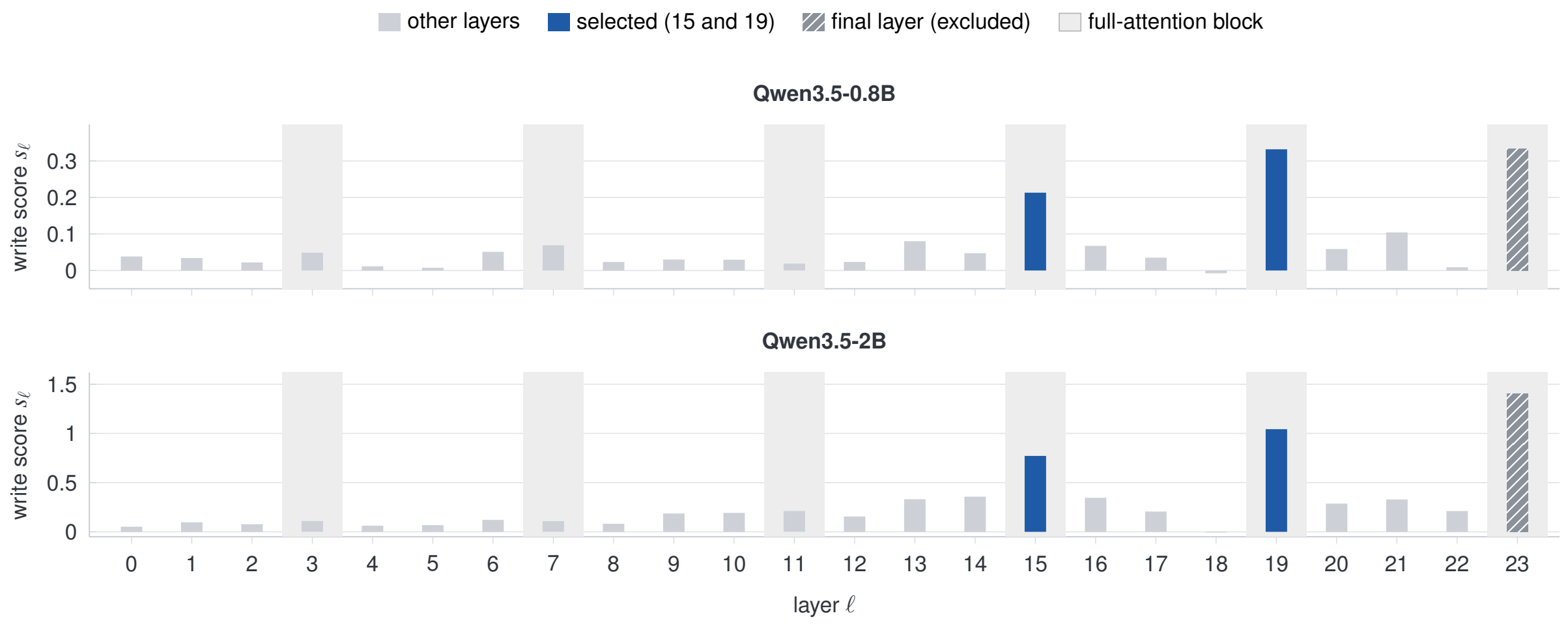
\begin{figure}[t]
\centering
\begin{tikzpicture}
\begin{groupplot}[group style={group size=1 by 2, vertical sep=0.85cm, xticklabels at=edge bottom},
    paperaxis, set layers, ybar, /pgf/bar width=6.2pt, width=\textwidth, height=3.25cm, xmin=-0.7, xmax=23.7, enlarge x limits=false,
    xtick={0,...,23}, ymin=-0.05, ylabel={write score $s_\ell$}, ylabel style={yshift=-2pt},
    legend style={at={(0.5,1)}, anchor=south, yshift=22pt, legend columns=4, /tikz/every even column/.append style={column sep=6pt}},
    legend image code/.code={\draw[#1, draw=none] (0cm,-0.08cm) rectangle (0.22cm,0.10cm);}]
  \nextgroupplot[ymax=0.40, ytick={0,0.1,0.2,0.3}, yticklabel style={/pgf/number format/fixed},
                 title={Qwen3.5-0.8B}]
    \pgfplotsinvokeforeach{3,7,11,15,19,23}{\fill[black!7, on layer=axis background] (axis cs:{#1-0.5},-0.05) rectangle (axis cs:{#1+0.5},0.40);}
    \addplot[fill=rule, draw=none, bar shift=0pt] coordinates {(0,0.0384) (1,0.0342) (2,0.0221) (3,0.0488) (4,0.0116) (5,0.0074)
      (6,0.0512) (7,0.0691) (8,0.0233) (9,0.0300) (10,0.0295) (11,0.0189) (12,0.0234) (13,0.0802) (14,0.0474) (16,0.0675)
      (17,0.0352) (18,-0.0077) (20,0.0588) (21,0.1043) (22,0.0089)};
    \addplot[fill=oursblue, draw=none, bar shift=0pt] coordinates {(15,0.2131) (19,0.3319)};
    \addplot[fill=basegray, draw=none, bar shift=0pt, postaction={pattern=north east lines, pattern color=white}] coordinates {(23,0.3337)};
    \addlegendimage{legend image code/.code={\draw[fill=black!7, draw=black!30, line width=0.3pt] (0cm,-0.08cm) rectangle (0.22cm,0.10cm);}}
    \legend{other layers, selected (15 and 19), final layer (excluded), full-attention block}
  \nextgroupplot[ymax=1.62, ytick={0,0.5,1,1.5}, yticklabel style={/pgf/number format/fixed}, xlabel={layer $\ell$},
                 title={Qwen3.5-2B}]
    \pgfplotsinvokeforeach{3,7,11,15,19,23}{\fill[black!7, on layer=axis background] (axis cs:{#1-0.5},-0.05) rectangle (axis cs:{#1+0.5},1.62);}
    \addplot[fill=rule, draw=none, bar shift=0pt] coordinates {(0,0.0519) (1,0.0968) (2,0.0773) (3,0.1098) (4,0.0626) (5,0.0679)
      (6,0.1214) (7,0.1089) (8,0.0818) (9,0.1864) (10,0.1925) (11,0.2117) (12,0.1558) (13,0.3310) (14,0.3574) (16,0.3457)
      (17,0.2061) (18,-0.0069) (20,0.2869) (21,0.3295) (22,0.2114)};
    \addplot[fill=oursblue, draw=none, bar shift=0pt] coordinates {(15,0.7718) (19,1.0434)};
    \addplot[fill=basegray, draw=none, bar shift=0pt, postaction={pattern=north east lines, pattern color=white}] coordinates {(23,1.4064)};
\end{groupplot}
\end{tikzpicture}
\caption{\textbf{Forward layer scan of both models.} Write score of every MLP along the loop direction, on loop tokens
relative to productive tokens (\cref{eq:scan}). Shaded columns mark full-attention blocks; all other blocks use linear
attention. The rule ``two highest layers, excluding the final layer'' was fixed on Qwen3.5-0.8B, where it returns the agent's
choice, and selects layers 15 and 19 again on Qwen3.5-2B. In both models the three highest layers are full-attention blocks,
so the agreement may follow from the shared architecture.}
\label{fig:layerscan}
\end{figure}

%% file: sections/04_experiments.tex
\section{Repairing Qwen3.5-0.8B: Comparison with the LLM Agent}
\label{sec:exp}

\subsection{Experimental Setup}
\label{sec:exp-setup}

\paragraph{Model and Edit.} We apply our method to the \code{down\_proj} matrices ($1{,}024\times3{,}584$) in layers 15 and 19 of Qwen3.5-0.8B~\citep{qwen35card}, matching the layers targeted by the agent edit. We configure the optimization with $K=512$ candidates per row, $\beta=0.5$, and a strict budget of 65,536 changed weights per layer. The target dose is set to 1.6 times the dose of the agent edit along the direction $u$, corresponding to targets of 6.7 and 6.1 times the layer's own write for layers 15 and 19, respectively (\cref{sec:step3}); consequently, the two edits are not strictly dose-matched. Notably, the forward scan detailed in \cref{sec:step2} selects these exact same two layers. However, the exclusion rule for this scan was finalized after observing the initial scores, and the entire recipe was designed retrospectively by analyzing the agent's public logs. We refer to this specific configuration as the \ours edit. Unless otherwise specified, this edit utilizes the simulated annealing solver; variants marked with "SB" denote the identical recipe solved via simulated bifurcation (\cref{sec:step5}). A complete list of hyperparameters and settings is provided in \cref{app:implementation}.

\paragraph{Data.} The forward statistics are derived from 384 practice problems released alongside the agent's evaluation harness, comprising 192 problems from the MATH training split and 192 from the GSM8K training split~\citep{hendrycks2021math,cobbe2021gsm8k}. An additional 128 held-out practice problems constitute the selection screen used for the ablation studies in \cref{sec:exp-ablations}. We note that early exploratory variants were evaluated on a fixed subset of 100 MATH-500 problems; consequently, the full MATH-500 results include problems encountered during the development phase, a limitation we address in \cref{sec:limitations}.

\paragraph{Evaluation.} We strictly adhere to the evaluation protocol established by the agent's authors~\citep{takehi2026logs}: generation is performed using vLLM~\citep{kwon2023vllm} in thinking mode, with a temperature of 0.6, top-$p$ of 0.95, top-$k$ of 20, a presence penalty of 1.5, and a maximum of 32,768 new tokens. The final boxed answer following the \code{</think>} token is graded using Math-Verify~\citep{mathverify}. We report performance across five benchmarks: MATH-500~\citep{lightman2024verify}, GSM8K~\citep{cobbe2021gsm8k}, AMC23~\citep{amc23}, Minerva Math~\citep{lewkowycz2022minerva}, and OlympiadBench~\citep{he2024olympiadbench}. The suffix @$k$ denotes the number of completions generated per problem, with accuracy defined as the mean correctness across these completions. To ensure a fair comparison, both edited checkpoints are evaluated within the identical harness using the same random seeds, and paired statistical tests are conducted on a problem-by-problem basis.

\subsection{Accuracy and Termination}
\label{sec:exp-accuracy}

\begin{table}[t]
\centering
\caption{\textbf{Benchmark accuracy on Qwen3.5-0.8B (\%).} The suffix @$k$ indicates the number of completions per problem. The upper block reproduces the numbers reported by \citet{takehi2026sparks}; the lower block evaluates the edited checkpoints within a unified harness, with \ours solved by simulated annealing (SA) or simulated bifurcation (SB). The bottom rows detail the percentage of completions cut at the 32,768-token cap.}
\label{tab:main}
\small
\begin{tabular}{@{}lcccccc@{}}
\toprule
  & MATH-500@1 & MATH-500@4 & GSM8K@1 & AMC23@16 & Minerva@4 & OlympiadBench@1 \\
\midrule
Base model, reported            & --   & 7.9  & 13.8 & 0.5  & 1.5  & 1.3 \\
Agent edit, reported            & --   & 48.1 & 67.1 & 23.4 & 25.4 & 21.8 \\
\midrule
Agent edit, our harness         & 51.2 & 51.7 & 65.6 & 22.0 & 24.2 & 21.4 \\
\oursrow \ours (SA)             & 54.0 & 54.6 & 70.5 & 22.7 & \textbf{26.1} & \textbf{24.3} \\
\oursrow \ours (SB)             & \textbf{54.4} & \textbf{55.1} & \textbf{70.8} & \textbf{25.0} & 25.6 & 24.0 \\
\midrule
\multicolumn{7}{@{}l}{\emph{Cut at cap (\%)}} \\
Agent edit, our harness         & 31.4 & 31.6 & 8.0  & 51.4 & 26.7 & 47.8 \\
\oursrow \ours (SA)             & 23.0 & \textbf{23.9} & \textbf{3.1} & 37.8 & \textbf{18.1} & 32.5 \\
\oursrow \ours (SB)             & \textbf{22.8} & \textbf{23.9} & 3.2 & \textbf{36.6} & 18.5 & \textbf{31.9} \\
\bottomrule
\end{tabular}
\par\smallskip
\begin{minipage}{\textwidth}\footnotesize
\textit{Paired comparisons against the agent edit.} \ours (SA): MATH-500@4 $+2.9\pm1.0$ points, $z=2.94$, $p\approx0.003$ (better on 108 problems, worse on 73); GSM8K: 155 problems solved exclusively by \ours versus 90 by the agent edit ($p=4\times10^{-5}$); OlympiadBench: 55 versus 35 ($p=0.045$); MATH-500@1: 57 versus 43 ($p=0.19$); AMC23: $+0.6\pm1.0$ (9 better, 7 worse); Minerva: $+1.9\pm1.2$ (47 better, 34 worse). \ours (SB): MATH-500@4 $+3.4\pm1.0$, $z=3.47$, $p\approx0.0005$ (113 better, 65 worse); GSM8K: 156 versus 87 ($p=10^{-5}$); OlympiadBench: 55 versus 37 ($p=0.08$); MATH-500@1: 52 versus 36 ($p=0.11$); AMC23: $+3.0\pm1.5$ (16 better, 7 worse); Minerva: $+1.5\pm1.2$ (46 better, 44 worse). The base model reaches the token cap on 93\% of MATH-500 completions. The agent's originally reported counts utilized two GSM8K and four OlympiadBench completions.
\end{minipage}
\end{table}

\Cref{tab:main} presents a direct comparison of the edits. Crucially, with either solver, \ours outperforms the agent edit across all five benchmarks. The performance gains on MATH-500 (with four completions per problem) and on GSM8K remain statistically significant even after applying a Bonferroni correction across all twelve comparisons with the agent edit (two solvers, six settings; $p<0.05/12$ in all four cases). Furthermore, with simulated annealing, the improvement on OlympiadBench is nominally significant ($p=0.045$), though it does not survive the strict correction. The remaining differences fall within the expected noise margin. 

Two key observations clarify the evaluation harness. First, \ours surpasses the performance metrics originally reported by the agent's authors on four of the five benchmarks, with the sole exception of AMC23. Second, our measured MATH-500 score for the agent edit (51.7\% accuracy with 31.6\% of answers cut) aligns closely with an independent reproduction of the reconstructed checkpoint on the full MATH-500 dataset, which reported 51.0\% accuracy with an identical 31.6\% cut rate~\citep{khaliq2026repro}. Consequently, the discrepancy with the originally reported 48.1\% appears to stem from the initial evaluation setup rather than our harness.

\paragraph{Problems Never Used During Development.} Early exploratory variants were screened on a fixed subset of 100 MATH-500 problems. On the remaining 400 problems, which were entirely unseen during development, \ours achieves 53.8\% accuracy compared to 50.8\% for the agent edit (with four completions per problem). This represents a paired difference of $+2.9\pm1.1$ points ($z=2.62$, $p\approx0.009$; 87 problems improved, 58 degraded), with the cut rate dropping from 31.2\% to 24.0\%. The SB edit scores 53.6\%, maintaining a $+2.8\pm1.1$ point advantage over the agent edit ($p\approx0.007$; 85 better, 52 worse). Notably, the 0.5-point lead of the SB edit over annealing on the full 500-problem set originates entirely from the 100 development problems; for annealing, the performance gap on the held-out problems matches the overall gap.

The most pronounced difference between the edits lies in their termination behavior. \ours reaches the token cap less frequently across every benchmark. For instance, on GSM8K, the cut rate plummets from 8.0\% to 3.1\%, and on OlympiadBench, it drops from 47.8\% to 32.5\%. These accuracy gains are primarily driven by the model's newfound ability to successfully conclude its reasoning within the allocated token budget.

\paragraph{Solver Choice.} Both solvers yield repairs of equivalent quality. Simulated bifurcation lands marginally closer to the target doses ($-2.189$ and $-3.083$ against targets of $-2.183$ and $-3.056$, whereas annealing reaches $-2.225$ and $-3.075$), while also exhibiting lower collateral in layer 15 and modifying 4\% fewer weights (108,327 versus 112,630; see \cref{tab:dose,tab:footprint}). Crucially, no benchmark statistically separates the two edits: on MATH-500@4, the SB edit leads by a negligible $+0.45\pm0.9$ points ($p=0.62$); on GSM8K, the two solvers uniquely solve 115 and 111 problems, respectively ($p=0.84$); and the largest observed gap, $+2.3\pm1.3$ points on AMC23, remains statistically insignificant. This demonstrates that the performance gains stem fundamentally from the mathematical formulation itself, rather than from the tuning of a specific solver.

\subsection{Cost of a Single Repair}
\label{sec:exp-cost}

\begin{table}[t]
\centering
\caption{\textbf{Computational cost of producing a single edit.} This comparison contrasts the agent's exhaustive discovery run against a single execution of our fixed recipe, which reuses the agent's targeted layers and dose. Agent figures are derived from the released 50-pass logs~\citep{takehi2026logs} at list prices~\citep{openai_pricing}; GPU time is priced at the median A100-40GB on-demand rate of \$1.76 per hour~\citep{amcompute2026}. A single forward-pass request accommodates up to 512 generations or 1,024 teacher-forced texts. Method development costs are excluded for both approaches.}
\label{tab:cost}
\small
\begin{tabularx}{\textwidth}{@{}l Y Y@{}}
\toprule
 & Agent Edit & \ours \\
\midrule
Search procedure      & LLM agent (GPT-6 Astra, maximal reasoning effort) & Simulated annealing on 2,048 row QUBOs \\
Agent turns and tokens & 158 turns, 89.7M tokens (80.3M cached input, 2.9M output) & None \\
Gradients             & None & None \\
Forward passes        & 50 & 4 (one generation, two statistics passes, one screen) \\
Model GPU time        & 12.2 h & 21.8 min \\
Search time           & 1.8 h of agent responses & 8.8 min of GPU annealing \\
Wall clock            & 15.6 h (20.3 h including a 4.7 h API outage) & \textbf{31 min} (30.6 min) \\
\midrule
LLM API cost          & \$291 (up to \$485 at long-context rates) & \$0 \\
GPU cost              & \$21 & \$0.90 \\
\textbf{Total}        & \textbf{\$312} & \textbf{\$0.90} \\
\bottomrule
\end{tabularx}
\end{table}

\Cref{tab:cost} quantifies the computational expense of a single repair. The agent expended 158 turns and nearly 90 million tokens, with its 50 forward-pass requests alone consuming 12.2 GPU-hours. In stark contrast, \ours requires only 4 forward-pass requests: one generation pass over the practice problems, two teacher-forced passes for statistics collection, and one validation screen. These operations consume merely 21.8 GPU-minutes, representing a 34-fold reduction in forward-pass compute compared to the agent, and the subsequent annealing adds just 8.8 minutes. Consequently, our repair is approximately 30 times faster and 350 times cheaper, while requiring zero LLM tokens. Notably, the agent's preliminary runs with 10 and 25 forward passes, which failed to yield a functional fix, incurred an additional \$270 at the same pricing tier.

It is critical to contextualize this comparison: it contrasts the exhaustive process of discovering a fix with the efficient application of a fixed recipe. The agent bore the computational burden of identifying the failure mode, the loop direction, and the target layers. Our recipe was designed retrospectively by analyzing its public logs, reusing its layer selections, and calibrating the dose based on its final edit. Both columns explicitly exclude method development costs. Furthermore, \cref{sec:transfer} demonstrates the application of this recipe to a novel model without relying on any of these prior dependencies.

\subsection{Edit Footprint}
\label{sec:exp-footprint}

\begin{table}[t]
\centering
\caption{\textbf{Weight and bit modifications across both edited matrices.} State-tree fields comprise the 8-bit shared node, the 4-bit state, the sign bit, and the 7-bit mantissa; modifying a shared node inherently alters the states of all other weights within its group. Neither edit alters escape membership, and changes confined to existing escape payloads are excluded from this count.}
\label{tab:footprint}
\small
\begin{tabular}{@{}lrrrrrrr@{}}
\toprule
 & \makecell[r]{Changed\\Weights} & \makecell[r]{BF16 Bit\\Flips} & \makecell[r]{State-Tree\\Bit Flips} & \makecell[r]{Sign\\Flips}
 & \makecell[r]{State\\Changes} & \makecell[r]{Shared\\Exponents} & \makecell[r]{Mantissa\\Changes} \\
\midrule
Agent Edit         & 130,188 & 621,388 & 720,586 & 19,299 & 146,981 & 1,261 & 129,668 \\
\oursrow \ours (SA) & 112,630 & 278,767 & 181,062 & 57,704 & 99,952 & \textbf{0} & \textbf{0} \\
\oursrow \ours (SB) & \textbf{108,327} & \textbf{268,090} & \textbf{174,245} & 55,666 & \textbf{96,007} & \textbf{0} & \textbf{0} \\
\bottomrule
\end{tabular}
\end{table}

\Cref{tab:footprint} details the precise structural modifications introduced by each edit. \ours modifies 13\% fewer weights (affecting only 1.5\% of the two matrices), reduces BF16 bit flips by a factor of 2.2, and cuts state-tree bit flips by a factor of four; the SB edit achieves an additional 4\% reduction in size. Crucially, while the agent edit rewrites the mantissa of nearly every weight it touches and shifts 1,261 shared exponents, \ours performs neither operation, a guarantee enforced by the construction of its admissible move set. The most frequently applied moves are scaling by $\times4$ and $\times(-4)$, followed by $\times2$, $\times(-2)$, and sign flips. Consequently, a patch of this nature is exceptionally compact, mathematically exact, and entirely native to the stored format.

\subsection{Ablation on the Objective and Continuous Edits}
\label{sec:exp-ablations}

\begin{table}[t]
\centering
\caption{\textbf{Ablation study on the practice screen} (128 held-out problems, one completion each; the standard error of accuracy is approximately 4.4 points). The selected configuration utilizes $\beta=0.5$, a $1.6\times$ dose, the eight admissible moves from \cref{eq:moves}, simulated annealing, and a budget of 65,536 changed weights per layer. Each row in the middle block isolates a single modified setting. The lower block solves the identical objective without sparsity or discrete moves, using the same direction $u$ and dose, and subsequently rounds the result to BF16; $\bar c$ denotes the mean diagonal of $C_{\Pset}$.}
\label{tab:ablations}
\small
\begin{tabular}{@{}llrrrr@{}}
\toprule
Variant & Change & Accuracy (\%) & Cut at cap (\%) & \makecell[r]{Changed\\Weights} & \makecell[r]{BF16 Bit\\Flips} \\
\midrule
\oursrow \ours & Selected configuration & \textbf{57.0} & 22.7 & 112.6k & 279k \\
Loop only               & $\beta=0$                     & 43.8 & 53.1 & 112.3k & 277k \\
Weaker conclude term    & $\beta=0.25$                  & 53.1 & 28.1 & 115.4k & 285k \\
Stronger conclude term  & $\beta=1$                     & 53.9 & 21.1 & 93.2k  & 232k \\
Lower dose              & $1.2\times$                   & 57.8 & 16.4 & 77.4k  & 193k \\
Higher dose             & $2.0\times$                   & 54.7 & 26.6 & 118.5k & 301k \\
Basic moves             & $f\in\{-1,2,\frac12\}$        & 50.0 & 22.7 & 122.7k & 179k \\
Greedy solver           & Zero temperature              & 53.9 & 18.0 & 112.0k & 277k \\
Simulated bifurcation   & SB in place of annealing      & 56.3 & 18.8 & 108.3k & 268k \\
Smaller budget          & 28k weights per layer         & 49.2 & 24.2 & 45.6k  & 121k \\
\midrule
Agent Edit              & Released checkpoint           & 52.3 & 25.0 & 130.2k & 621k \\
\midrule
\multicolumn{6}{@{}l}{\emph{Continuous edits of the same objective}} \\
Closed-form rank-one    & $\Delta\propto-u\,\bigl((C_{\Pset}+10^{-3}\bar c I)^{-1}\mu_{\R}\bigr)^{\top}$ & 57.8 & 12.5 & 7.02M & 28.7M \\
Stronger ridge          & $10^{-1}\bar c$ in place of $10^{-3}\bar c$ & 56.3 & 7.0 & 6.94M & 26.6M \\
Directional ablation    & $\Delta=-\alpha\,u\,u^{\top}W$ & 38.3 & 25.0 & 6.92M & 26.1M \\
\bottomrule
\end{tabular}
\end{table}

\Cref{tab:ablations} systematically varies one parameter at a time on the practice screen. The most pronounced effect arises from the inclusion of the concluding direction. Without it ($\beta=0$), the edit successfully suppresses the loop, but the model still fails to terminate, resulting in accuracy stagnating in the mid-40s with 53\% of answers hitting the token cap. Incorporating this term, however, elevates the accuracy well above that of the agent edit. Similarly, restricting the admissible moves to merely sign flips and single doublings or halvings, or severely shrinking the weight budget, degrades performance. Doses ranging from $1.2\times$ to $2.0\times$ yield comparable results, and both annealing and simulated bifurcation outperform a greedy solver, although the screen lacks the resolution to distinguish differences of a few percentage points. The $1.6\times$ dose was fixed during the initial development phase, prior to this dose sweep, which was subsequently executed around the selected configuration. While we did not reselect the configuration post-sweep, it is worth noting that the lower dose achieves a marginally higher score while modifying 31\% fewer weights. No strictly dose-matched ($1.0\times$) edit was evaluated.

\paragraph{Continuous Edits.} If we remove the sparsity penalty and the discrete move constraints, \cref{eq:energy} admits a closed-form minimizer: the rank-one update $\Delta=-\tfrac{\gamma}{2}\,u\,(C_{\Pset}^{-1}\mu_{\R})^{\top}$, which structurally mirrors the updates employed by ROME~\citep{meng2022rome}. When rounded to BF16 at an equivalent dose, this continuous solution scores 57.8\% on the screen, matching \ours within the screen's margin of error, and reaches the token cap even less frequently. However, it fundamentally alters 96\% of the weights across both matrices (7.0 million changes versus 112,630) and induces 103 times as many BF16 bit flips. A plain directional ablation~\citep{arditi2024refusal} at the same dose scores a mere 38.3\%. These continuous variants were evaluated exclusively on the screening set, not on the full benchmarks. Crucially, they demonstrate that the mathematical objective itself, rather than the discrete solver, drives the majority of the accuracy gains. Furthermore, they highlight that \ours achieves the accuracy of the optimal continuous solution while modifying 62 times fewer weights, with all changes written natively into the stored format.

\paragraph{Which Size Matters.} We quantify the number of changed stored weights because the edit is applied as a direct patch to the stored checkpoint: the repaired model is served using ordinary weights, in place, with an entirely unchanged inference graph. From a purely descriptive standpoint, the rank-one edit appears smaller, requiring only two vectors (4,608 numbers per layer) compared to the approximately 56,000 (position, move) pairs per layer required by \ours. However, this compactness holds only if the edit is preserved as a separate rank-one adapter during inference. When written directly into the weights, the continuous edit rewrites 96\% of the stored values and nearly every mantissa, necessitating the complete re-encoding of any compressed, deduplicated, or state-tree checkpoint. In stark contrast, a \ours patch leaves every unrelated stored bit entirely untouched, with each modification confined strictly to a single sign bit or a 4-bit state.

%% file: sections/05_transfer.tex
\section{Transfer to Qwen3.5-2B Without a Reference Edit}
\label{sec:transfer}

In our evaluation of Qwen3.5-0.8B, \ours relied on the agent's selected layers and calibrated its dose based on the agent's final edit. In this section, we eliminate both dependencies. We apply the method to a novel model, autonomously selecting its target layers via the forward scan and calibrating the dose using the base-model rule defined in \cref{sec:step3}, while leaving the remainder of the recipe entirely unchanged.

\paragraph{The Failure Mode.} Qwen3.5-2B~\citep{qwen35card2b} comprises 24 layers with \code{down\_proj} matrices of dimensions $2{,}048\times6{,}144$. Under greedy decoding (temperature 0, thinking mode, 8,192-token cap), the base model correctly answers only 21.6\% of the 1,319 GSM8K test problems (\cref{tab:transfer}). The underlying cause mirrors the failure observed in the smaller model: 77.9\% of its generations reach the token cap, and 1,018 of these 1,028 truncated answers exhibit looping behavior. Crucially, when the model does successfully terminate, it provides the correct answer 97.9\% of the time. This indicates that the model possesses the requisite problem-solving capability but fails to halt its reasoning. We specifically selected greedy decoding for this evaluation because an initial probe on 200 GSM8K test problems revealed that the base model struggles most severely under these conditions; consequently, the recipe has not yet been evaluated under the model's recommended sampling settings or on the MATH-500 benchmark.

\paragraph{The Recipe.} We first prompt the base model to answer the same 384 practice problems using greedy decoding, yielding a 16.9\% success rate with 83.1\% of the answers truncated. The forward scan subsequently ranks layers 23, 19, 15, and 14 highest (\cref{fig:layerscan}, bottom). Applying our exclusion rule for the final layer isolates layers 15 and 19, precisely matching the pair selected for the smaller model. The forward statistics are derived from 275 looping and 65 correct traces. We set the target dose to $6.5$ times each layer's own write, and allocate a budget of 200,000 changed weights per layer, representing approximately 1.6\% of a single matrix and closely aligning with the 1.8\% utilized for the smaller model. All remaining hyperparameters are identical to those detailed in \cref{sec:exp}. The entire pipeline, encompassing the layer scan and a single validation screen, executes in just 26 minutes on a single A100 GPU. Crucially, this process requires no LLM agent, no reference checkpoint, and no gradient computations.

\input{figures/fig_transfer}

\begin{table}[t]
\centering
\caption{\textbf{Performance of Qwen3.5-2B on GSM8K.} Evaluated across all 1,319 test problems using greedy decoding in thinking mode with an 8,192-token cap. The \ours recipe utilizes the scan-selected layers and the base-model dose rule without any modifications. Paired counts detail the number of problems solved exclusively by the edit, exclusively by the base model, and by both.}
\label{tab:transfer}
\small
\begin{tabular}{@{}lccccccr@{}}
\toprule
Model & Layers & Dose & Accuracy (\%) & Cut at cap (\%) & \makecell{Only edit / only base\\/ both} & \makecell[r]{Changed\\weights} & \makecell[r]{State-tree\\bit flips} \\
\midrule
Qwen3.5-2B                      & --         & --          & 21.6 & 77.9 & --            & --      & -- \\
\oursrow \ours recipe           & 15, 19     & $6.5\times$ & \textbf{69.4} & 26.7 & 669 / 38 / 247 & 292,845 & 473,524 \\
\ours, stronger dose            & 14, 15, 19 & $9\times$   & \textbf{71.8} & 24.4 & 687 / 25 / 260 & 482,423 & 788,492 \\
\bottomrule
\end{tabular}
\par\smallskip
\begin{minipage}{\textwidth}\footnotesize
\textit{Statistical comparisons against the base model:} McNemar $p<10^{-100}$ for both edits. \textit{Stronger dose versus the standard recipe:} 171 problems solved exclusively by the stronger edit and 140 exclusively by the recipe ($p=0.09$). Crucially, neither edit alters a shared exponent, a mantissa, or escape membership. Total BF16 bit flips: 718,546 for the standard recipe and 1,178,678 for the stronger dose.
\end{minipage}
\end{table}

\paragraph{Results.} Applying the unchanged recipe elevates GSM8K accuracy from 21.6\% to 69.4\%, representing a substantial gain of 47.8 percentage points (\cref{tab:transfer}). The edit successfully solves 669 problems that the base model fails to answer, while incorrectly altering the solution for 38 of the 285 problems the base model originally solved correctly (a 13\% degradation on those specific instances). Notably, when evaluated on the 1,119 problems outside the initial 200-problem probe used to select the decoding setting, the performance remains consistent: 69.6\% accuracy versus 21.6\% for the base model (571 problems solved exclusively by the edit, 34 exclusively by the base model, and 208 by both). Furthermore, the proportion of truncated answers drops precipitously from 77.9\% to 26.7\%. While the accuracy on fully completed answers decreases slightly from 97.9\% to 94.7\%, it is important to note that these figures refer to different subsets of problems; the 38 lost problems confirm that the edit does perturb some correct base-model trajectories. Structurally, the patch modifies a mere 1.2\% of the weights across the two matrices and, consistent with our design, alters no shared exponents and no mantissas.

\paragraph{Dose Calibration.} Reasoning loops prove more persistent under greedy decoding than under sampling, and the practice screen indicates that performance continues to improve with higher doses (\cref{fig:transfer}, right). Consequently, we also evaluated a dose of $9\times$ alongside the inclusion of the next highest-scoring scan layer, layer 14. The best screened variant, targeting layers 14, 15, and 19 at a $9\times$ dose, achieves 71.8\% on GSM8K, a 50.2-point improvement over the base model. However, its advantage over the standard recipe is not statistically significant, and it incurs a cost of 1.7 times as many state-tree bit flips. Therefore, we report the standard recipe as our primary result. Crucially, the majority of this performance gain stems from the dose rule and the layer selection mechanism, both of which were established prior to evaluating this specific model.

%% file: figures/fig_transfer.tex
\begin{figure}[t]
\centering
\begin{tikzpicture}
\begin{groupplot}[group style={group size=2 by 1, horizontal sep=1.55cm}, paperaxis, height=4.6cm]
  \nextgroupplot[width=7.4cm, ybar, /pgf/bar width=15pt, ymin=0, ymax=112, ytick={0,25,50,75,100},
                 symbolic x coords={base,recipe,strong}, xtick=data, enlarge x limits=0.24,
                 xticklabels={Qwen3.5-2B, {\ours recipe}, {stronger dose}},
                 ylabel={share of GSM8K problems (\%)},
                 title={GSM8K test, 1,319 problems, greedy decoding},
                 nodes near coords, every node near coord/.append style={font=\sffamily\scriptsize, text=ink},
                 legend style={at={(0.99,0.995)}, anchor=north east, legend columns=2, /tikz/every even column/.append style={column sep=5pt}},
                 legend image code/.code={\draw[#1] (0cm,-0.08cm) rectangle (0.22cm,0.10cm);}]
    \addplot[fill=oursblue, draw=oursblue] coordinates {(base,21.6) (recipe,69.4) (strong,71.8)};
    \addplot[fill=agentfill, draw=agentorange, postaction={pattern=north east lines, pattern color=agentorange!70}]
             coordinates {(base,77.9) (recipe,26.7) (strong,24.4)};
    \legend{accuracy, cut at cap}
  \nextgroupplot[width=7.0cm, xmin=-0.4, xmax=9.6, xtick={0,4,6.5,9}, xticklabels={0,4,6.5,9},
                 ymin=0, ymax=114, ytick={0,25,50,75,100}, xlabel={dose ($\times$ own write)}, ylabel={share of screen problems (\%)},
                 title={Practice screen, 128 problems, greedy decoding},
                 legend style={at={(0.99,0.985)}, anchor=north east, font=\sffamily\tiny, legend columns=2, row sep=-1.5pt, /tikz/every even column/.append style={column sep=4pt}, fill=white, fill opacity=0.9, text opacity=1},
                 enlarge x limits=false]
    \addplot[oursblue, line width=1.0pt, mark=*, mark size=1.7pt] coordinates {(0,13.3) (4,32.8) (6.5,45.3) (9,53.1)};
    \addplot[oursblue!55, line width=1.0pt, mark=square*, mark size=1.6pt] coordinates {(0,13.3) (6.5,43.0) (9,54.7)};
    \addplot[agentorange, line width=0.9pt, dashed, mark=*, mark size=1.5pt, mark options={solid}] coordinates {(0,86.7) (4,65.6) (6.5,53.1) (9,44.5)};
    \addplot[agentorange!55, line width=0.9pt, dashed, mark=square*, mark size=1.4pt, mark options={solid}] coordinates {(0,86.7) (6.5,56.2) (9,42.2)};
    \legend{{accuracy, 2 layers},{accuracy, 3 layers},{cut, 2 layers},{cut, 3 layers}}
    \draw[softink, dotted, line width=0.6pt] (axis cs:6.5,0) -- (axis cs:6.5,100);
    \node[font=\sffamily\tiny, text=softink, anchor=south west, rotate=90] at (axis cs:6.32,3) {recipe dose};
\end{groupplot}
\end{tikzpicture}
\caption{\textbf{The recipe transfers to Qwen3.5-2B.} Left: on all 1,319 GSM8K problems the unchanged recipe (layers 15 and
19, dose $6.5\times$) lifts accuracy from 21.6\% to 69.4\% and cuts the share of looping answers by two thirds; layers 14, 15
and 19 at $9\times$ reach 71.8\%. Right: on the practice screen, accuracy rises and the cut share falls steadily with dose, both for layers 15 and 19 and for layers 14, 15 and 19.}
\label{fig:transfer}
\end{figure}
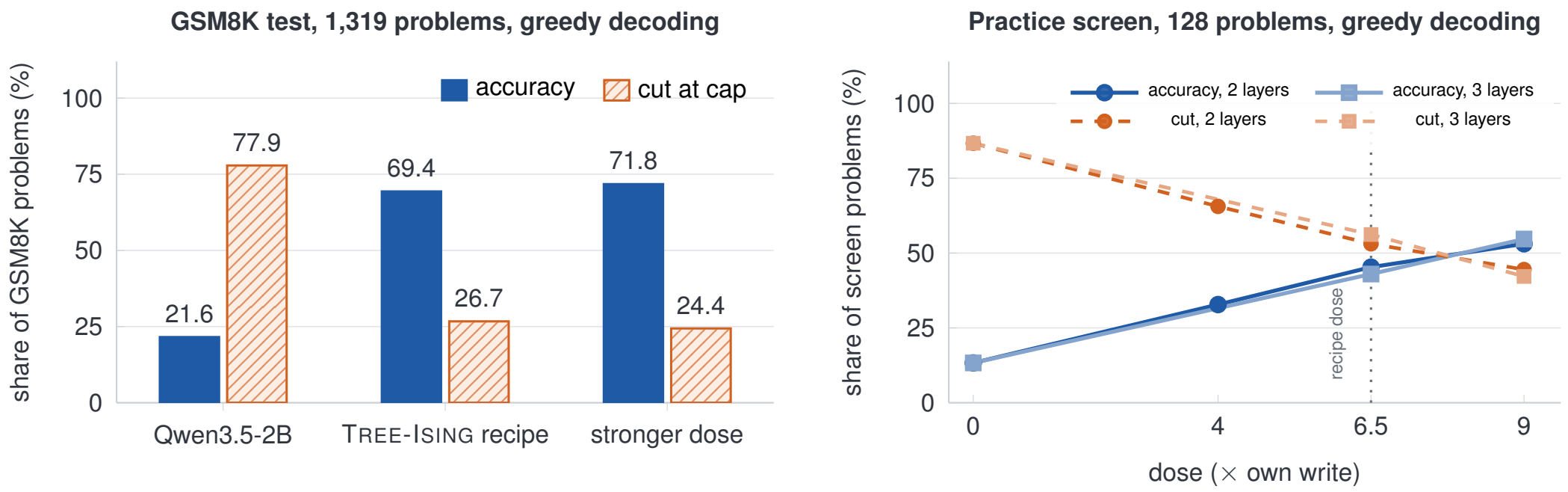

%% file: sections/06_quantum.tex
\section{Quantum-Hardware Proof of Concept}
\label{sec:quantum}
\label{sec:q-ibm}

We report a small proof of concept for executing a repair-derived Ising objective on an IBM Heron quantum processor using the quantum approximate optimization algorithm (QAOA)~\citep{farhi2014qaoa,zhou2020qaoa}. This specific instance isolates eight candidate weights from a single row of layer 15 in Qwen3.5-0.8B, with all other variables in the row fixed to zero. The separately calibrated parameter $\gamma=0.0844$ adheres to a single-weight threshold rule, and we set $\lambda=\lambda_0$ (\cref{app:hardware}). With eight binary variables yielding $2^8=256$ possible assignments, exhaustive enumeration establishes the exact optimum at negligible computational cost.

We optimized the QAOA angles classically using exact statevector simulation. In a single hardware execution, the depth-two QAOA circuit sampled the optimum in 110 out of 1,024 shots, compared to merely four shots for a uniform control. We note that uniform sampling serves as a relatively weak baseline for assessing optimization performance; matched random-angle circuits and classical solvers applied to identical instances would provide a more rigorous comparison. Full circuit configurations, sampling results, and resource accounting are detailed in \cref{app:hardware}.

Crucially, this experiment establishes that a small, separately calibrated instance can be successfully encoded, executed, and sampled on quantum hardware. However, it does not constitute evidence of a quantum advantage. As demonstrated in \cref{sec:exp}, simulated annealing, simulated bifurcation, and the closed-form continuous edit all yield statistically indistinguishable accuracy, indicating that the current row-level problems are not computationally challenging for classical solvers, nor do marginal improvements in row energy translate to superior model performance. The primary value of these problems for quantum optimization lies in their role as a source of structured, real-world Ising instances equipped with a meaningful downstream metric: the accuracy of the repaired model. Significantly harder formulations, such as joint move selection or explicit bit budgets (\cref{app:extensions}), would be required before a quantum solver could offer a tangible advantage. Consequently, the full-model accuracy results reported in \cref{sec:exp,sec:transfer} are derived entirely from the classical GPU solver. A subsequent investigation into hybrid refinement would condition each block on the GPU solution, apply the established repair calibration, and evaluate the resulting checkpoint on standard benchmarks.

%% file: sections/07_related.tex
\section{Related work}
\label{sec:related}

\paragraph{Weight-space programming and model editing.}
\citet{takehi2026sparks} investigated whether a general-purpose LLM can read and write network weights; among other tasks, their agent successfully repairs the reasoning loops of Qwen3.5-0.8B. We study the identical repair task but replace the LLM agent with an optimizer built from the same empirical evidence. Classical model editing typically modifies factual associations through targeted updates of MLP weights~\citep{meng2022rome,meng2023memit}, locates knowledge within individual neurons~\citep{dai2022knowledge}, learns editors that predict weight updates~\citep{mitchell2022mend}, or combines fine-tuning differences~\citep{ilharco2023task}. Crucially, these methods produce continuous updates. In contrast, \ours selects sparse, discrete changes that the stored weight format can represent exactly, and \cref{sec:exp-ablations} compares our approach with a ROME-like rank-one solution of our own objective. Furthermore, while bit-flip attacks demonstrate that a few flipped bits can substantially alter a network's behavior~\citep{rakin2019bfa}, we leverage this same sensitivity to repair rather than to harm.

\paragraph{Activation steering and directional interventions.}
Steering methods typically add directions to the residual stream at inference time~\citep{li2023iti,turner2023steering,zou2023repe}, and directional ablation can be folded directly into the weights~\citep{arditi2024refusal}. Our directions are contrasts of the same kind, measured at the onset of loops and at the conclusion of successful thinking traces. However, our intervention takes the form of a sparse weight edit optimized against a collateral penalty, requiring no modifications at inference time. Mechanistic studies of transformer circuits strongly motivate reading such directions directly from the residual stream~\citep{elhage2021framework}. Closest to our setting, ThinkEdit~\citep{sun2025thinkedit} identifies a linear direction governing reasoning length, isolates a small set of attention heads driving overly short reasoning, and removes that direction from their output projections, altering approximately 0.2\% of the parameters. We address the opposite failure mode---reasoning that never terminates---and edit MLP output projections through sparse discrete moves rather than a continuous projection.

\paragraph{Repetition and looping.}
Degenerate repetition has been extensively addressed through decoding heuristics~\citep{holtzman2020curious} and modified training objectives~\citep{welleck2020unlikelihood,xu2022loop}, and recent reports on reasoning models analyze how extended thinking traces interact with decoding configurations~\citep{deepseek2025r1,qwen35card}. In contrast, \ours alters neither the decoder nor the training process; it simply edits the weights once, entirely offline.

\paragraph{Weight formats.}
Block floating-point and microscaling formats share exponents within groups of weights~\citep{drumond2018hbfp,rouhani2023mx}, and lossless BF16 compression exploits the low entropy of these exponents~\citep{zhang2025dfloat}. State trees~\citep{ali2026everyweight} provide a general representation of neural networks as paths of shared discrete decisions, with lossless BF16 storage and low-bit compression serving as specific instances; that work also heals a compressed model using gradient and curvature estimates. In our formulation, the state tree fundamentally defines the admissible moves of a forward-only objective.

\paragraph{QUBO formulations of network weights.}
AdaRound derives a binary rounding problem from a second-order approximation of the task loss~\citep{nagel2020adaround}, and \citet{muniz2025qubo} formulate layer-output reconstruction for quantization as an exact QUBO that separates into independent output rows, solving it via simulated annealing. Layer-output reconstruction also underpins many pruning and quantization methods~\citep{hassibi1993obs,frantar2023gptq}. Our objective augments this reconstruction term with a behavioral term along a measured direction and a sparsity cost, operating over moves that preserve the stored format intact. Formulated as Ising models, these row problems connect seamlessly to the broad toolbox of Ising formulations~\citep{lucas2014ising}, simulated annealing~\citep{kirkpatrick1983sa}, quantum annealing~\citep{kadowaki1998qa,johnson2011qa}, QAOA~\citep{farhi2014qaoa}, and digital or analogue Ising machines~\citep{goto2019sb,mohseni2022ising}.

\paragraph{Quantum hardware for neural networks and LLMs.}
Quantum annealers have been utilized to train binary neural networks~\citep{sasdelli2021bnn} and to select pruning and quantization configurations for convolutional networks~\citep{wang2025aqc}. For LLMs, recent work integrates variational quantum circuits into a pretrained model, either as parameter-efficient fine-tuning modules executed on a superconducting processor~\citep{kong2025qtha} or as unitary adapters inside the projection layers of Llama 3.1 8B running on IBM hardware~\citep{aizpurua2026cayley}. In those studies, the quantum circuit constitutes a novel addition to the model architecture. In \ours, however, the model form remains entirely unchanged: the quantum device, much like the GPU, simply solves for which of the model's existing BF16 weights should be modified, as demonstrated here on a single eight-variable subproblem.

%% file: sections/08_limitations.tex
\section{Limitations}
\label{sec:limitations}

\paragraph{Scope.} Our study addresses a single failure mode: reasoning loops, specifically within two models of a single family that share a hybrid linear/full-attention architecture, evaluated exclusively on mathematical benchmarks. We have not assessed instruction-following capabilities, although \citet{takehi2026sparks} report that the agent edit raised IFEval from 21.0\% to 62.3\%, nor have we tested general language modeling or non-mathematical benchmarks. Consequently, side effects remain unmeasured beyond the collateral term, which protects the layer's output only on tokens derived from the practice problems. Evaluating larger models, alternative model families, and other behavioral failure modes remains a direction for future work.

\paragraph{Baselines and Evaluation.} The base model's 7.9\% accuracy on MATH-500 is the figure reported by \citet{takehi2026sparks}; we did not independently rerun the base model in our evaluation harness. Furthermore, the continuous edits detailed in \cref{sec:exp-ablations} were evaluated exclusively on the 128-problem practice screen. Early exploratory variants on Qwen3.5-0.8B were screened on 100 MATH-500 problems, meaning the full MATH-500 results include problems exposed during development; however, the performance gap remains consistent on the 400 strictly held-out problems (\cref{sec:exp-accuracy}). Each benchmark evaluation utilized a single generation seed. The $p$-values reported in the tables are nominal; notably, only the gains on MATH-500@4 and GSM8K over the agent edit survive a strict Bonferroni correction across the twelve agent comparisons.

\paragraph{Design Choices.} Our repair recipe was formulated retrospectively after analyzing the agent's public logs. On Qwen3.5-0.8B, the edit utilizes the agent's selected layers and a dose 1.6 times that of the agent's, calibrated on the practice screen; notably, no strictly dose-matched edit was executed. While the Qwen3.5-2B experiment eliminates both of these dependencies, the rule excluding the final layer was established after inspecting the 0.8B scan. Furthermore, the layers selected in both models are full-attention blocks, suggesting that this agreement may stem from their shared architectural design rather than independent validation. For Qwen3.5-2B, greedy decoding and the GSM8K benchmark were selected based on base-model results from a 200-problem GSM8K test subset where the base model struggles most, whereas all edit-specific settings were determined solely on the practice problems.

\paragraph{Surrogate Objective and Solvers.} Our objective functions as a first-order surrogate derived from the statistics of the unedited model: it inherently ignores how an edit in one layer alters the inputs to subsequent layers, and the dose serves merely as a linear proxy for the desired behavioral shift. What remains mathematically exact is the decomposition of this surrogate into independent row problems once candidate columns and a single move per weight are fixed; however, these two choices, along with the row solvers themselves, are heuristic. The practice screen resolves performance differences of approximately four percentage points, leaving the optimal dose and move set somewhat uncertain. Each solver was executed only once with a single seed, and their run times were not measured under strictly matched conditions; therefore, we make no definitive claims regarding their relative computational speed.

\paragraph{Accounting.} Our state-tree bit counts assume a straightforward encoder and deliberately exclude modifications within the payloads of existing escapes; furthermore, direct patching of a compressed checkpoint has not yet been implemented. The reported time and cost metrics cover a single execution of a fixed recipe at list prices and explicitly exclude method development costs on both sides. Stage timings are derived from single runs executed with two versions of the solver script and should not be interpreted as a rigorous scaling measurement.

\paragraph{Quantum Hardware.} The quantum hardware evaluation is confined to a single eight-variable subproblem, encompassing just 8 of the 7.3 million weights across the two edited matrices, with the remainder of its row fixed to zero. This evaluation utilized a dose calibration declared separately from the published edit, angles optimized classically, a single hardware job, and one calibration period. Crucially, the resulting solution was not written back into a model. The control baseline sampled uniformly; a random-angle control on the identical compiled circuits has not been executed. Consequently, this result demonstrates a correct mapping and objective-consistent sampling on quantum hardware, not a computational advantage over classical solvers. Furthermore, \cref{eq:amdahl} bounds the theoretical benefit of any faster solver within the confines of our current recipe.

%% file: sections/09_conclusion.tex
\section{Conclusion}
\label{sec:conclusion}

While a frontier LLM can repair a looping model by directly editing its weights, we have demonstrated that this same repair task can be formulated as a sparse discrete optimization problem and solved in a matter of minutes. Specifically, \ours restricts admissible edits to sign flips and exponent shifts within the state tree of each weight group, evaluates them using forward statistics, and efficiently solves one small Ising problem per output row. On Qwen3.5-0.8B, our method produces an edit that is more accurate than the agent's baseline while modifying fewer weights and far fewer bits, all for a computational cost of approximately \$0.90 per run. Crucially, both simulated annealing and the quantum-inspired simulated bifurcation solver yield repairs of equivalent quality, demonstrating that the performance gains stem fundamentally from the mathematical formulation itself rather than the tuning of a specific solver. In contrast, a dense closed-form solution of the identical objective achieves comparable accuracy only by altering nearly every weight across both matrices. Furthermore, when applied to Qwen3.5-2B—with target layers and dosage calibrated entirely from the base model—this same recipe elevates GSM8K accuracy under greedy decoding from 21.6\% to 69.4\%.

Finally, an eight-variable QAOA experiment executed on IBM quantum hardware serves as a preliminary proof of concept for running a repair-derived Ising instance on a quantum processor. Extending this validation to full-layer repairs and rigorously measuring the resulting downstream model accuracy remains a compelling direction for future work.

\paragraph{Code Availability.}
The source code implementing \ours will be made publicly available shortly at \url{https://github.com/TheQuantumGOAT/Tree-Ising}.

%% file: sections/A_implementation.tex
\section{Implementation details}
\label{app:implementation}

\begin{table}[p]
\centering
\caption{Experimental settings for the two model repairs. Reported stage runtimes were measured on a single NVIDIA A100 40\,GB GPU.}
\label{tab:impl}
\footnotesize
\setlength{\tabcolsep}{4pt}
\renewcommand{\arraystretch}{1.08}

\begin{tabularx}{\textwidth}{@{}p{0.21\textwidth} Y Y@{}}
\toprule
Setting
 & Qwen3.5-0.8B (\cref{sec:exp})
 & Qwen3.5-2B (\cref{sec:transfer}) \\
\midrule

\multicolumn{3}{@{}l}{\textbf{Edit configuration}} \\
\addlinespace[3pt]

Target tensors
 & \code{down\_proj}, layers 15 and 19
 & \code{down\_proj}, layers 15 and 19.
   The stronger variant also includes layer 14 \\

Tensor dimensions
 & $1{,}024\times3{,}584$ per layer
 & $2{,}048\times6{,}144$ per layer \\

Layer selection
 & Agent-selected layers, corroborated by a forward scan
 & Forward scan, excluding the final layer \\

Candidate neurons
 & $K=512$ of 3,584, with the same candidate columns across all rows within a layer
 & $K=512$ of 6,144 per row \\

Allowed moves
 & Eight multiplicative factors from \cref{eq:moves}
 & Same \\

Edit direction
 & $u\propto\uloop-0.5\,\uend$, computed per layer
 & Same \\

Dose target
 & $1.6\times$ the agent edit's dose along $u$.
   Equivalent to $6.7\times$ and $6.1\times$ the model's own write
   in layers 15 and 19, respectively
 & $6.5\times$ the model's own write.
   Stronger variant uses $9\times$ \\

Edit budget
 & 65,536 changed weights per layer
 & 200,000 changed weights.
   Stronger variant permits 300,000 \\

\midrule
\multicolumn{3}{@{}l}{\textbf{Statistics collection}} \\
\addlinespace[3pt]

Collection data
 & 384 practice problems, comprising 192 MATH-train and
   192 GSM8K-train problems, with one completion each
 & Same problems \\

Decoding
 & Agent protocol, temperature 0.6
 & Greedy \\

Token cap
 & 8,192
 & 8,192 \\

Collected traces
 & 224 looping, 33 correct
 & 275 looping, 65 correct \\

Tokens per layer
 & $|\R|=195$k, $|\Pset|=567$k,
   $|\Zset|=2.1$k, $|\Mset|=120$k
 & $|\R|=246$k, $|\Pset|=655$k,
   $|\Zset|=4.2$k, $|\Mset|=127$k \\

\midrule
\multicolumn{3}{@{}l}{\textbf{Optimization}} \\
\addlinespace[3pt]

Simulated annealing
 & 16 replicas and 300 sweeps.
   Temperature decreases from 2 to $10^{-4}\times$ the median $|h|$.
   Calibration uses 4 replicas and 60 sweeps
 & Same \\

Simulated bifurcation
 & Discrete SB with 1,200 steps of size 0.5.
   Pump ramp over the first 25\% of steps, global field scaling,
   and best-state tracking every 20 steps.
   Followed by float64 single-bit descent with at most 1,024 flips.
   Batches of 32 rows, with 16 replicas for the final solve
   and 4 for calibration
 & Not run \\

\midrule
\multicolumn{3}{@{}l}{\textbf{Selection and evaluation}} \\
\addlinespace[3pt]

Selection data
 & 128 held-out practice problems,
   comprising 96 MATH and 32 GSM8K problems
 & Same problems \\

Selection decoding
 & One completion per problem, agent protocol, 32k-token cap
 & One completion per problem, greedy, 8k-token cap \\

Evaluation protocol
 & Shared protocol described below
 & Shared protocol, with greedy decoding and an 8k-token cap \\

\midrule
\multicolumn{3}{@{}l}{\textbf{Stage runtime (minutes)}} \\
\addlinespace[3pt]

Generation          & 5.7  & 5.9 \\
Layer scan          & ---  & 3.6 \\
Statistics          & 5.1  & 4.8 \\
Simulated annealing & 8.8  & 6.7 \\
Selection screen    & 11.0 & 5.0 \\

\bottomrule
\end{tabularx}

\smallskip
\begin{minipage}{\textwidth}
\footnotesize
\textbf{Shared evaluation protocol.}
Evaluation uses vLLM 0.32 with BF16 precision and thinking enabled.
Each problem is followed by
``Please reason step by step, and put your final answer within
\textbackslash boxed\{\}.''
The last boxed answer after \code{</think>} is graded using Math-Verify.
\end{minipage}
\end{table}

\begin{table}[h]
\centering
\caption{Dose calibration of every edited layer. Own write is $|u^{\top}W\mu_{\R}|$ of the base model; doses are
$u^{\top}\Delta\mu_{\R}$ after rounding to BF16. $\lambda$ attempts counts the values of $\lambda$ tried, with a full $\gamma$
search at each.}
\label{tab:dose}
\small
\begin{tabular}{@{}llrrrrrcr@{}}
\toprule
Model & Layer & \makecell[r]{Own\\write} & \makecell[r]{Agent\\dose} & \makecell[r]{Target\\dose} & \makecell[r]{Achieved\\dose} & \makecell[r]{Achieved\\/ own} & \makecell{$\lambda$\\attempts} & \makecell[r]{Changed\\weights} \\
\midrule
Qwen3.5-0.8B & 15 & 0.328 & $-1.364$ & $-2.183$  & $-2.225$  & 6.8 & 1 & 51,754 \\
Qwen3.5-0.8B & 19 & 0.498 & $-1.910$ & $-3.056$  & $-3.075$  & 6.2 & 1 & 60,876 \\
Qwen3.5-0.8B, SB & 15 & 0.328 & $-1.364$ & $-2.183$  & $-2.189$  & 6.7 & 1 & 48,860 \\
Qwen3.5-0.8B, SB & 19 & 0.498 & $-1.910$ & $-3.056$  & $-3.083$  & 6.2 & 1 & 59,467 \\
Qwen3.5-2B   & 15 & 1.035 & --       & $-6.730$  & $-6.884$  & 6.6 & 1 & 111,211 \\
Qwen3.5-2B   & 19 & 1.755 & --       & $-11.405$ & $-11.522$ & 6.6 & 2 & 181,634 \\
\bottomrule
\end{tabular}
\end{table}

Every layer reaches at least its target dose, overshooting it by 0.3 to 2.3\% (SB by 0.3 and 0.9\%), and only layer 19 of Qwen3.5-2B needed a second value of $\lambda$ to meet its budget. The two annealing times in \cref{tab:impl} come from different versions of the solver
script, the second a generic reimplementation used for Qwen3.5-2B, run at different times; they include model loading and a $\gamma$ search whose length varies, and should not be read as a scaling measurement.

\paragraph{Hardware interpretation.} Each row objective has the Ising form used by quantum annealers and other specialized optimization machines. \Cref{sec:q-blocks,sec:q-protocol} describe the conditional-block route to hardware and the evaluation protocol. All
reported checkpoints were built on one GPU, with simulated annealing or, where marked SB, with simulated bifurcation.

%% file: sections/B_hardware.tex
\section{Quantum hardware experiment details}
\label{app:hardware}
\sloppy

\subsection{Instance and calibration}
The block statistics were computed from 224 looping traces and 33 correct traces. They include 195,013 loop tokens and 567,053 protected tokens per layer, together with 2,112 concluding and 120,068 continuing tokens from the correct traces. We downloaded only the layer-15 and layer-19 \code{down\_proj} tensors from the public Qwen3.5-0.8B checkpoint (revision \code{2fc0636}); their byte ranges and hashes are documented in the reproducibility package. We retained $K=512$, the eight moves in \cref{eq:moves}, $\beta=0.5$, and the sparsity scale
\begin{equation}
  \lambda=2\,\operatorname{median}_{r,a,\,f\in\{-1,2,\frac12\}}\;C_{\Pset,j_aj_a}\bigl[(f-1)W_{rj_a}\bigr]^2 .
\end{equation}
Since the calibrated $\gamma$ used for the published edit was not saved with the statistics, we specified a separate calibration rule before examining any solution. For each candidate weight, we computed the single-weight entry threshold
\begin{equation}
  t_{ra}=\min_{f\ \text{admissible},\ \ell_{raf}<0}\ \frac{C_{\Pset,j_aj_a}\,\delta_{raf}^2+\lambda}{-\ell_{raf}},\qquad
  \ell_{raf}=u_r\,\mu_{\R,j_a}\,\delta_{raf},
\end{equation}
which marks the point at which activating that weight alone becomes energetically favorable. We set $\gamma$ to $(1+10^{-6})$ times the 65,536th smallest of the 505,622 finite positive thresholds, obtaining
\[\gamma=0.0844304006,\qquad \lambda=5.88178264\times10^{-7}.\]

We then selected layer 15 and the row with the largest $|u_r|$ (row 5), restricting the block to its first eight candidate columns,
$(0, 46, 1184, 294, 371, 504, 264, 1130)$, while fixing all other variables in that row to zero. The block was constructed in float64 from the exact BF16 weights. For all 256 assignments, its energy agrees with a direct evaluation of the dose, collateral, and sparsity terms to within $7.05\times10^{-19}$. Exhaustive search identifies the unique optimum
\begin{equation}
  x^{\star}=(1,1,0,0,1,0,1,1),\qquad \mathcal{E}_B(x^{\star})=-7.68966437\times10^{-4}.
\end{equation}
Qiskit displays this bitstring as \code{11010011}, reflecting its ordering of the bits as $x_7\dots x_0$.
\subsection{Circuits}

Under Qiskit's convention $Z_i=1-2x_i$, the cost Hamiltonian associated with $\mathcal{E}_B(x)=h^{\top}x+x^{\top}Jx$ is
\begin{equation}
  \widehat H_C=\sum_i\Bigl(-\frac{h_i}{2}-\frac12\sum_j J_{ij}\Bigr)Z_i+\sum_{i<j}\frac{J_{ij}}{2}Z_iZ_j,\qquad
  E_0=\frac12\sum_i h_i+\frac14\sum_{ij}J_{ij},
\end{equation}
where adding $E_0$ to each eigenvalue recovers $\mathcal{E}_B$ for all 256 assignments. Defining $D=\mathrm{diag}(\mathcal{E}_B(x)/s)$
with $s=7.32314592\times10^{-5}$ and using the mixer $B=\sum_jX_j$, we write the depth-$p$ state as
\begin{equation}
  |\psi_p(\bm\vartheta,\bm\varphi)\rangle=e^{-i\varphi_{p-1}B}e^{-i\vartheta_{p-1}D}\cdots e^{-i\varphi_0B}e^{-i\vartheta_0D}\,|+\rangle^{\otimes8}.
\end{equation}
We optimize the angles by minimizing the exact statevector mean $\langle\psi_p|D|\psi_p\rangle$ using L-BFGS-B with analytic
adjoint gradients~\citep{jones2020adjoint}, subject to $\vartheta_\ell\in[0,2\pi]$ and $\varphi_\ell\in[0,\pi]$. At each depth,
we use 48 starts, including states inherited from the preceding depth, split cost and mixer layers, schedules interpolated
in the spirit of parameter transfer~\citep{zhou2020qaoa}, perturbed continuations, and random angles. Each start is limited
to 800 iterations and 4,000 evaluations, with seed 17. Retaining the best state encountered ensures that the ideal mean
energy does not increase with depth.

We restricted the live IBM Fez target to the eight physical qubits used in the first run. At each depth, we tried four
compiler seeds (17 to 20) at optimization level 3 with exact synthesis. We retained the compilation with the fewest CZ
gates, breaking ties by selecting the smallest circuit depth. Gate twirling, measurement twirling, and dynamical decoupling
were disabled. Each QAOA circuit was paired with a uniform-sampling control measured using the same final readout assignment.

\begin{table}[h]
\centering
\caption{Device and execution settings. Circuit statistics are taken from the archived submitted circuits.}
\label{tab:device}
\small
\begin{tabularx}{\textwidth}{@{}l Y@{}}
\toprule
Backend     & \code{ibm\_fez}, IBM Heron r2, 156 fixed-frequency transmon qubits with tunable couplers and heavy-hexagonal connectivity~\citep{ibm_hardware,ibm_processors,ibm_technology} \\
Access      & IBM Quantum Platform Open Plan, \code{us-east}, job mode, using Qiskit 2.6.2 and IBM runtime client 0.51.0 \\
Physical qubits & 121, 122, 123, 124, 125, 136, 142, 143 \\
Problem     & 8 binary variables and 28 nonzero logical couplings \\
Shots       & 1,024 per circuit, with six circuits (three QAOA depths and their controls) submitted in one job \\
Compiled QAOA circuits & $p=1$ has 107 CZ gates and depth 211, $p=2$ has 220 CZ gates and depth 387, and $p=3$ has 350 CZ gates and depth 617 \\
Reference device graph & \code{conf\_fez.json} from qiskit-ibm-runtime 0.51.0, backend version 1.2.8, with 176 undirected connections~\citep{ibm_fez_snapshot} \\
\bottomrule
\end{tabularx}
\end{table}

\begin{table}[h]
\centering
\caption{Physical qubit assignments. Classical bit $j$ records variable $x_j$. Each control uses the final readout assignment of its corresponding QAOA circuit.}
\label{tab:mapping}
\small
\begin{tabular}{@{}lcccccccc@{}}
\toprule
Variable & $x_0$ & $x_1$ & $x_2$ & $x_3$ & $x_4$ & $x_5$ & $x_6$ & $x_7$ \\
\midrule
Circuit entry      & 123 & 124 & 122 & 136 & 125 & 121 & 143 & 142 \\
Readout, $p=1$     & 142 & 143 & 121 & 122 & 136 & 123 & 125 & 124 \\
Readout, $p=2$     & 125 & 121 & 124 & 122 & 136 & 123 & 142 & 143 \\
Readout, $p=3$     & 142 & 143 & 121 & 122 & 136 & 125 & 123 & 124 \\
\bottomrule
\end{tabular}
\end{table}

\begin{table}[h]
\centering
\caption{Classically optimized angles in radians, listed in increasing layer order. Evaluation counts and times are cumulative,
as optimization at each depth builds on the preceding depth.}
\label{tab:angles}
\small
\begin{tabular}{@{}clrrr@{}}
\toprule
$p$ & Cost angles $\bm\vartheta$ & Mixer angles $\bm\varphi$ & Evaluations & Time (s) \\
\midrule
1 & (0.474617)                   & (2.772206)                   & 769   & 0.115 \\
2 & (0.462255, 0.650097)         & (2.599518, 2.662857)         & 1,975 & 0.430 \\
3 & (0.342703, 0.641411, 0.645390) & (2.654418, 2.681259, 2.847049) & 3,496 & 1.011 \\
\bottomrule
\end{tabular}
\end{table}
\subsection{Hardware sampling results}

\input{figures/fig_qpu}

\begin{table}[!htbp]
\centering
\caption{\textbf{Eight-variable QAOA proof of concept on IBM Heron} (\code{ibm\_fez}, one job, 1,024 shots per circuit). The optimum rate is the fraction of
samples matching the unique exact optimum, reported with a binomial Wilson 95\% interval. Ideal rates are exact statevector
probabilities at the same angles. Energies are multiplied by $10^4$. Lower values are better, and the optimum is $-7.690$.}
\label{tab:qpu}
\small
\begin{tabular}{@{}lrrrrr@{}}
\toprule
Circuit & Native CZ & Optimum hits & Optimum rate (\%) [95\% interval] & Ideal rate (\%) & Mean energy $\pm$ SE \\
\midrule
QAOA, $p=1$            & 107 & 33  & 3.22 [2.30, 4.49]   & 2.31  & $-5.617\pm0.040$ \\
\rowcolor{qfill} QAOA, $p=2$ & 220 & \textbf{110} & \textbf{10.74} [8.99, 12.79] & 20.33 & $\mathbf{-5.859\pm0.047}$ \\
QAOA, $p=3$            & 350 & 55  & 5.37 [4.15, 6.93]   & 37.06 & $-5.756\pm0.046$ \\
\midrule
Uniform control, $p=1$ & 0   & 4   & 0.39 [0.15, 1.00]   & 0.39  & $-4.476\pm0.048$ \\
Uniform control, $p=2$ & 0   & 4   & 0.39 [0.15, 1.00]   & 0.39  & $-4.454\pm0.050$ \\
Uniform control, $p=3$ & 0   & 7   & 0.68 [0.33, 1.40]   & 0.39  & $-4.448\pm0.047$ \\
\bottomrule
\end{tabular}
\end{table}

The observed optimum rate was highest at depth two (10.74\%), followed by depth three (5.37\%) and depth one (3.22\%)
(\cref{tab:qpu}). The ideal statevector probability of the optimum increases with depth, whereas the measured probability
peaks at depth two. These measurements were obtained in a single job during one calibration period. The uniform controls
contain zero native CZ gates, compared with 107 to 350 in the QAOA circuits. Matched circuits with random angles would
provide a stronger control for assessing the effect of the optimized angles.
\subsection{Resources and an earlier diagnostic}

The exact ideal mean energies at the three depths are $(-5.636507,-6.727090,-7.120713)\times10^{-4}$, compared with the minimum
$-7.689664\times10^{-4}$. IBM reports 4 seconds of QPU time for the six circuits and 6,144 shots (job
\code{dbouna8ou4as73csnpd0}, 9 November 2026). According to the service timestamps, 2,648.9 seconds elapsed between creation
and completion, including 2,352.7 seconds before the job entered the running state. Queueing therefore accounts for most
of the elapsed time. Classical preparation took 61 seconds in total.

An earlier single-circuit diagnostic (job \code{dborrfs4148s739tgkig}) evaluated the same block at $p=1$ under a different
compilation policy, yielding 119 CZ gates and depth 254. The optimum was sampled 21 times in 1,024 shots (2.05\%, Wilson
interval [1.35, 3.12]\%), compared with 6 times (0.59\%, [0.27, 1.27]\%) for its uniform control. The corresponding ideal
rates were 2.31\% and 0.39\%, and the reported QPU time was 3 seconds. The edge counts in \cref{fig:qpu} correspond to the
119 CZ gates from this earlier run. The package accompanying the paper includes the QUBO, calibration metadata, compiled
circuits, raw counts, job identifiers, input hashes, and scripts.

\subsection{Prospective extension to larger repair blocks}
\label{sec:q-blocks}

A dense row of $K=512$ variables contains $K(K-1)/2=130{,}816$ pairwise couplings. Minor embedding requires a physical
connection for each logical coupling, but the working graph of D-Wave's \code{Advantage2\_system1} lists only 41,205 couplers
in total~\citep{dwave_props}, ruling out an embedding of the full row. Dense embeddings on current annealers accommodate at
most a few hundred variables~\citep{dwave_embedding,dwave_samplers}. A future hardware backend could therefore operate on
conditional blocks. Let $\mathcal{E}_r(x)=h^{\top}x+x^{\top}Jx$, where $J$ is symmetric with a zero diagonal. For a variable
block $B$, fixing its complement $F$ at the current solution $x_F$ gives the block problem
\begin{equation}
  \mathcal{E}_B(z)=z^{\top}J_{BB}\,z+\bigl(h_B+2J_{BF}x_F\bigr)^{\top}z+\mathrm{const}
  \label{eq:block}
\end{equation}
which is again a QUBO. Its effective field incorporates the influence of every fixed variable. A hybrid loop would start
from the GPU solution, submit difficult blocks to the quantum device, and accept updates only when they decrease the full
row energy. The classical component would handle the calibration of $\gamma$ and $\lambda$. Submission counts also matter.
A single pass over disjoint blocks of size $b$ requires $2{,}048\lceil 512/b\rceil$ submissions, amounting to 16,384 at $b=64$,
before any repeated sweeps or calibration.

\subsection{Comparison protocol for future hardware extensions}
\label{sec:q-protocol}

Hardware and classical solvers should be evaluated on identical, archived instances, using the same statistics, candidate
moves, penalties, blocks, fixed assignments, and starting states. Reports should include the achieved energy, its gap to
the best known or certified optimum, feasibility, coefficient precision, chain breaks, and the complete time budget.
Timing should distinguish programming, annealing, and readout~\citep{dwave_timing} from embedding, data transfer, and queueing.
The end-to-end evaluation should measure the repaired checkpoint's accuracy on fresh problems.

\begin{samepage}
The stage times in \cref{tab:cost} also limit the speedup available from a faster solver. The recipe requires 21.8 minutes
of model computation and 8.8 minutes of solver time. If the solver took no time at all, the maximum speedup would be
\begin{equation}
  S_{\max}=\frac{21.8+8.8}{21.8}\approx1.40 .
  \label{eq:amdahl}
\end{equation}
Assessing a future hardware extension would require matched-instance comparisons of both solution quality and end-to-end
performance.
\end{samepage}

%% file: figures/fig_qpu.tex
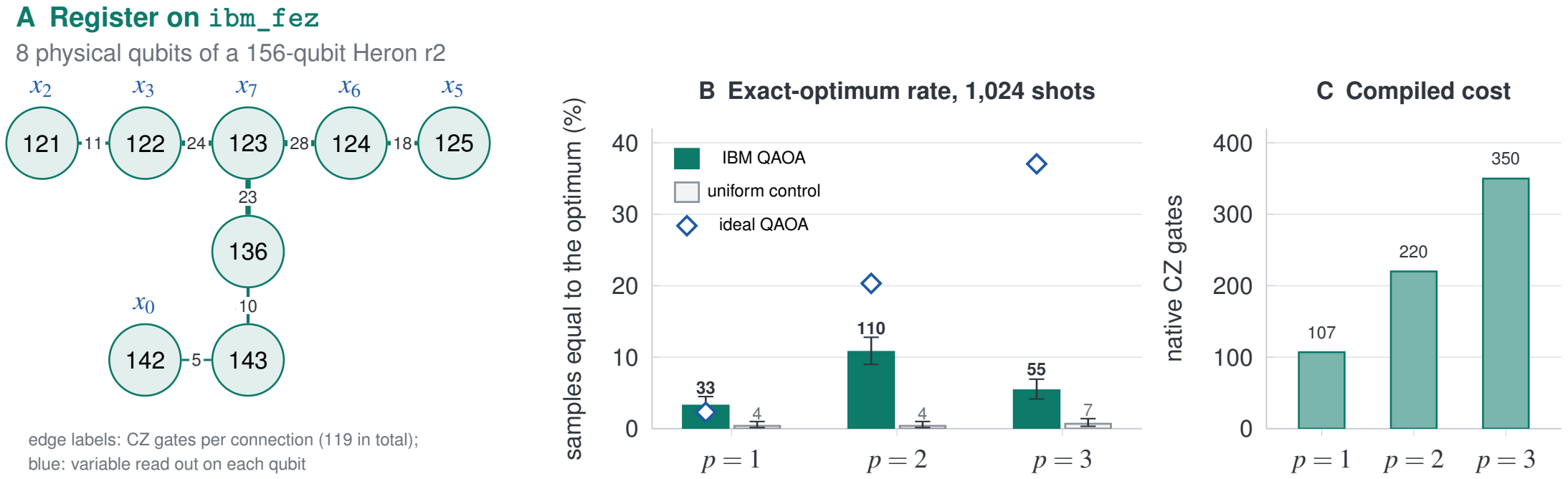
\begin{figure}[t]
\centering
\begin{tikzpicture}[x=1cm,y=1cm]
  \begin{scope}[shift={(0,0)}]
    \node[panelhead=qteal] at (-0.25,3.62) {A\enspace Register on \code{ibm\_fez}};
    \node[panelsub] at (-0.25,3.28) {8 physical qubits of a 156-qubit Heron r2};
    \foreach \q/\x/\y/\v in {121/0/2.35/x_2, 122/0.95/2.35/x_3, 123/1.90/2.35/x_7, 124/2.85/2.35/x_6, 125/3.80/2.35/x_5,
                             136/1.90/1.35/x_4, 143/1.90/0.35/x_1, 142/0.95/0.35/x_0} {
      \node[qnode=qteal, minimum size=0.66cm, font=\sffamily\scriptsize] (q\q) at (\x,\y) {\q};
      \node[font=\sffamily\scriptsize, text=oursblue] at ($(q\q)+(0.0,0.50)$) {$\v$};
    }
    \foreach \a/\b/\n in {121/122/11, 122/123/24, 123/124/28, 124/125/18, 123/136/23, 136/143/10, 142/143/5} {
      \draw[qteal, line width={0.35pt+0.045pt*\n}] (q\a) -- node[font=\sffamily\tiny, text=ink, fill=white, inner sep=0.6pt] {\n} (q\b);
    }
    \node[font=\sffamily\tiny, text=softink, align=left, anchor=north west] at (-0.25,-0.20)
      {edge labels: CZ gates per connection (119 in total);\\ blue: variable read out on each qubit};
  \end{scope}
\end{tikzpicture}\hspace{0.55cm}%
\begin{tikzpicture}
\begin{groupplot}[group style={group size=2 by 1, horizontal sep=1.15cm}, paperaxis, height=4.35cm]
  \nextgroupplot[width=6.1cm, ybar=1.5pt, /pgf/bar width=12pt, ymin=0, ymax=42, ytick={0,10,20,30,40},
                 symbolic x coords={p1,p2,p3}, xtick=data, xticklabels={$p=1$,$p=2$,$p=3$}, enlarge x limits=0.24,
                 ylabel={samples equal to the optimum (\%)}, title={B\enspace Exact-optimum rate, 1,024 shots},
                 legend style={at={(0.02,0.98)}, anchor=north west, font=\sffamily\tiny},
                 legend image code/.code={\draw[#1] (0cm,-0.08cm) rectangle (0.22cm,0.10cm);}]
    \addplot[fill=qteal, draw=qteal, error bars/.cd, y dir=both, y explicit, error bar style={ink, line width=0.5pt}]
      coordinates {(p1,3.223) += (0,1.268) -= (0,0.919)  (p2,10.742) += (0,2.046) -= (0,1.752)  (p3,5.371) += (0,1.555) -= (0,1.221)};
    \addplot[fill=basefill, draw=basegray, error bars/.cd, y dir=both, y explicit, error bar style={ink, line width=0.5pt}]
      coordinates {(p1,0.391) += (0,0.609) -= (0,0.239)  (p2,0.391) += (0,0.609) -= (0,0.239)  (p3,0.684) += (0,0.720) -= (0,0.352)};
    \addlegendimage{legend image code/.code={\node[diamond, draw=oursblue, fill=white, line width=0.7pt, minimum size=5.2pt, inner sep=0pt] at (0.11cm,0.01cm) {};}}
    \legend{IBM QAOA, uniform control, ideal QAOA}
    \foreach \c/\v in {p1/2.305, p2/20.328, p3/37.059} {
      \edef\temp{\noexpand\node[diamond, draw=oursblue, fill=white, line width=0.7pt, minimum size=5.2pt, inner sep=0pt]
        at ([xshift=-6.75pt]axis cs:\c,\v) {};}\temp}
    \node[font=\sffamily\tiny\bfseries, text=ink] at ([xshift=-6.75pt]axis cs:p1,5.7) {33};
    \node[font=\sffamily\tiny\bfseries, text=ink] at ([xshift=-6.75pt]axis cs:p2,14.1) {110};
    \node[font=\sffamily\tiny\bfseries, text=ink] at ([xshift=-6.75pt]axis cs:p3,8.3) {55};
    \node[font=\sffamily\tiny, text=softink] at ([xshift=6.75pt]axis cs:p1,2.1) {4};
    \node[font=\sffamily\tiny, text=softink] at ([xshift=6.75pt]axis cs:p2,2.1) {4};
    \node[font=\sffamily\tiny, text=softink] at ([xshift=6.75pt]axis cs:p3,2.6) {7};
  \nextgroupplot[width=4.3cm, ybar, /pgf/bar width=12pt, ymin=0, ymax=420, ytick={0,100,200,300,400},
                 symbolic x coords={p1,p2,p3}, xtick=data, xticklabels={$p=1$,$p=2$,$p=3$}, enlarge x limits=0.30,
                 ylabel={native CZ gates}, title={C\enspace Compiled cost},
                 nodes near coords, every node near coord/.append style={font=\sffamily\tiny, text=ink}]
    \addplot[fill=qteal!55, draw=qteal] coordinates {(p1,107) (p2,220) (p3,350)};
\end{groupplot}
\end{tikzpicture}
\caption{\textbf{A statistics-derived repair block on IBM quantum hardware.} (A) The eight-variable block runs on eight connected qubits of \code{ibm\_fez}; edge labels are the CZ counts of the earlier
$p=1$ diagnostic (119 CZ in total, \cref{app:hardware}), not of the 107-CZ circuit in (B) and (C). (B) Share of the 1,024 shots that equal the unique exact optimum, with Wilson 95\%
intervals, for QAOA at depths one to three and for matched uniform controls; diamonds give the ideal noiseless rate.
(C) Two-qubit gate count after compilation. Depth two gives the best measured rate; at depth three gate noise outweighs the
higher ideal rate.}
\label{fig:qpu}
\end{figure}

%% file: sections/C_extensions.tex
\section{Formulation extensions}
\label{app:extensions}

The results in this paper use the single-move formulation of \cref{sec:step4}. This appendix records three extensions
that enlarge the optimization problem, and with it the potential role of specialized solvers.

\paragraph{Joint choice among moves.} The current formulation preselects one move per candidate weight before solving the
row. Interactions can make a different move preferable once neighbouring weights change. Instead, let $i=(j,m)$ index a
candidate weight $j$ and an admissible move $m$, with $\delta_i=(f_m-1)W_{rj}$ and $y_i\in\{0,1\}$. At most one move per
weight may be active,
\begin{equation}
  \textstyle\sum_{m}y_{jm}\le 1 ,
\end{equation}
which a QUBO can enforce with a pairwise penalty $\rho\sum_j\sum_{m<m'}y_{jm}y_{jm'}$:
\begin{equation}
  \widetilde{\mathcal{E}}_r(y)=\sum_i A_iy_i+\sum_{i<k}B_{ik}y_iy_k+\rho\sum_j\sum_{m<m'}y_{jm}y_{jm'},
\end{equation}
with
\begin{equation}
  A_i=\gamma u_r\mu_{\R,j_i}\delta_i+C_{\Pset,j_ij_i}\delta_i^2+\lambda+\eta\,b_i,\qquad
  B_{ik}=2\,\delta_i\delta_k\,C_{\Pset,j_ij_k}.
\end{equation}
Here $b_i$ is the bit cost of move $i$ and $\eta\ge0$ prices it. Any $\rho>\max_i\bigl(|A_i|+\sum_{k\neq i}|B_{ik}|\bigr)$
makes every violating assignment suboptimal, because removing one conflicting move lowers the penalty by at least $\rho$ and
changes the rest of the objective by less. Such a bound widens the coefficient range, which matters for analogue hardware;
a categorical solver can keep one move or none per weight by construction instead. With eight moves the row grows to at
most $8K$ variables.

\paragraph{Bit costs.} For non-escaped weights whose moves keep the group maximum, the mantissa and the escape list, the
state-tree cost is additive across weights,
\begin{equation}
  b_i^{\mathrm{tree}}=\mathbf{1}\bigl[s_i'\neq s_i\bigr]+\operatorname{popcount}_4\bigl(D_i'\oplus D_i\bigr),
\end{equation}
where $s_i$ is the sign bit and $\oplus$ is bitwise exclusive or. Pricing these costs lets one budget written bits instead of
changed weights. Moves that change a group's shared exponent can be included by charging the node and the states of the
other weights in the group they shift; such moves are much more expensive in stored bits than tree-local ones.

\paragraph{Dose rules and budgets.} Global dose or cardinality penalties couple the rows. Linear Lagrange penalties with an
outer calibration loop, or explicit per-row budgets enforced by a categorical search, keep the rows independent. A sequential
variant can refresh forward statistics after editing one layer, so that the next layer's objective sees the edited inputs, at
the price of extra forward passes. Changes to the formulation and to the solver should be evaluated separately: first compare
solvers on identical single-move instances, then compare formulations at matched runtime and edit budgets, and finally
measure the repaired models on fresh problems and on unrelated tasks.